\documentclass{article}
\usepackage{proceed2e}
\usepackage[margin=1in]{geometry}
\usepackage{times}
\usepackage{amsmath}

\usepackage{amssymb}
\usepackage{bbm}
\usepackage{graphics}
\usepackage{graphicx}
\usepackage{booktabs}
\usepackage{caption}
\usepackage{subcaption}
\usepackage{color}
\usepackage{algorithm}
\usepackage{algorithmic}
\usepackage[round, comma, sort]{natbib}

\newcommand{\mvec}[1]{\mathbf{#1}}
\newcommand{\Trans}{{^{\mathrm{T}}}}

\title{Entropic Spectral Learning for Large-Scale Graphs}
\newtheorem{proposition}{Proposition}
\author{ {\bf Diego Granziol \thanks{\hspace{5pt} These two authors contributed equally} \hspace{0.1pt} \thanks{ \hspace{5pt} Machine Learning Research Group and Oxford-Man Institute for Quantitative Finance, Department of Engineering Science, University of Oxford}} , 
\hspace{5pt}
{\bf Binxin Ru \footnotemark[1]  \hspace{0.1pt} \footnotemark[2]} 
, \hspace{5pt}
{\bf Stefan Zohren\footnotemark[2]}  
, \hspace{5pt}
{\bf Xiaowen Dong\footnotemark[2]}  
, \hspace{5pt}
{\bf Michael Osborne\footnotemark[2]} 
, \hspace{5pt}
{\bf Stephen Roberts\footnotemark[2]} 
}
\begin{document}
\maketitle

\begin{abstract}
Graph spectra have been successfully used to classify network types, compute the similarity between graphs, and determine the number of communities in a network. For large graphs, where an eigen-decomposition is infeasible, iterative moment matched approximations to the spectra and kernel smoothing are typically used. We show that the underlying moment information is lost when using kernel smoothing. We further propose a spectral density approximation based on the method of Maximum Entropy, for which we develop a new algorithm. This method matches moments exactly and is everywhere positive. We demonstrate its effectiveness and superiority over existing approaches in learning graph spectra, via experiments on both synthetic networks, such as the Erd\H{o}s-R\'{e}nyi and Barab\'{a}si-Albert random graphs, and real-world networks, such as the social networks for Orkut, YouTube, and Amazon from the SNAP dataset.
\end{abstract}


\section{Introduction: Graphs and their Importance}\label{sec:intro}

Many systems of interest can be naturally characterised by complex networks; examples include social networks \citep{mislove2007measurement,flake2000efficient,leskovec2007dynamics}, biological networks \citep{palla2005uncovering} and technological networks. The biological cell can be compactly described as a complex network of chemical reactions. Trends, opinions and ideologies spread on a social network, in which people are nodes and edges represent relationships. The World Wide Web is a complex network of documents with web pages representing nodes and hyper-links denoting edges. Neural networks, considered state of the art machine learning algorithms for a variety of complex problems, can be seen as directed networks where neurons are the nodes and the synaptic connections between them are the edges. A variety of complex networks have been studied in the literature, from scientific collaborations \cite{ding2011scientific}, ecological/cellular networks \cite{fath2007ecological}, to sexual contacts \citep{albert2002statistical}. For a comprehensive introduction we refer the reader to \citep{newman2010networks}.

\subsection{Network spectra and applications}\label{subsec:net_spec_and_app}

Networks are mathematically represented by graphs. Of crucial importance to the understanding of the properties of a network or graph is its spectrum, which is defined as the eigenvalues of its adjacency or Laplacian matrix \citep{farkas2001spectra,cohen2018approximating}.
The spectrum  of a graph can be considered as a natural set of graph invariants and has been extensively studied in the fields of chemistry, physics and mathematics \cite{biggs1976graph}. Spectral techniques have been extensively used to characterise the global network structure \citep{newman2006modularity} and in practical applications thereof, such as facial recognition and computer vision \citep{belkin2003Laplacian}, learning dynamical thresholds \citep{mcgraw2008Laplacian}, clustering \cite{von2007tutorial}, and measuring graph similarity \citep{takahashi2012discriminating}. Applications in quantum chemistry include calculating the electron energy levels in hydrocarbons and their stability \cite{cvetkovic2009applications}, or minimising the energies of Hamiltonian systems in quantum physics \cite{stevanovic2011applications}. 
In the fields of statistics, computer science and machine learning, spectral clustering \citep{von2007tutorial} has become a powerful tool for grouping data, regularly outperforming or enhancing other classical algorithms, such as $k$-means or single linkage clustering. For most clustering algorithms, estimating the number of clusters is an open problem \citep{von2007tutorial}, with likelihood, ad-hoc, information theoretic, stability, and spectral approaches advocated. To this end, in spectral clustering, one analyzes the spectral gap in eigenvalues which we refer to as eigengap for short. An accurate estimate of the graph spectrum is therefore critical in this case, which will be discussed later in our paper.

\subsection{Problem statement}

A major limitation in utilizing graph spectra to solve interesting problems such as computing graph similarity and estimating the number of clusters is the inability to learn an everywhere-positive non-singular approximation to the spectral density in an automated and consistent fashion. Current methods rely on either full eigen-decompositions, which becomes prohibitive for large graphs, or iterative moment-matched approximations, both of which give a weighted Dirac sum that must be smoothed to be everywhere positive. Beyond requiring a choice of smoothing kernel $k_\sigma(x,x')$ and kernel bandwidth choice $\sigma$, or number of histogram bins, which are usually chosen in an ad-hoc manner, we show in this paper that for any smoothing kernel, the spectral moments, which can be seen to be representative of the underlying stochastic process and hence informative, are in fact biased away from their true values. In a nutshell, in order to make certain problems tractable, e.g., comparisons of network spectra, current methods loose the only exact information we have about the network. In this paper, we are interested in an efficient and accurate everywhere-positive approximation of the spectral density of large graphs.

\subsection{Main contributions}\label{subsec:contributions}

The main contributions of this paper are as follows:

\begin{itemize}
    \item We show that the method of kernel smoothing, commonly used in methods to estimate the spectral density, loses exact moment information;
    \item We propose a computationally efficient smooth spectral density approximation, based on the method of Maximum Entropy, which does not require kernel smoothing. It also admits analytic forms for symmetric and non-symmetric KL-divergences and Shannon entropy; 
    \item We show that our method is able to learn the underlying stochastic process of a network, and can be utilized for computing the similarity among networks from a wide range of synthetic and real world datasets;
    \item  We study the behaviour of bounds on changes in the graph spectrum upon perturbation of the graph, and its implication on determining the number of node clusters in the graph. 
    We further demonstrate the superior empirical performance of our method in learning the number of clusters compared to that of the Lanczos algorithm.
\end{itemize}


\section{Preliminaries} \label{sec: preliminaries}

Graphs are the mathematical structure underpinning the formulation of networks. Let $G = (V,E)$ be an undirected graph with vertex set $V = \{v_{i}\}_{i=1}^n$. Each edge between two vertices $v_{i}$ and $v_{j}$ carries a non-negative weight $w_{ij}>0$. $w_{ij}=0$ corresponds to two disconnected nodes. For un-weighted graphs we set $w_{ij}=1$ for two connected nodes. The \emph{adjacency matrix} is defined as $\mathbf{W}$ and  $w_{ij}=[\mathbf{W} ]_{ij}$. The degree of a vertex $v_{i} \in V$ is defined as 
\begin{equation} 
    d_{i} = \sum_{j=1}^{n}w_{ij}. 
\end{equation}
The \emph{degree matrix} $\mathbf{D}$ is defined as a diagonal matrix that contains the degrees of the vertices along diagonal, i.e., $\mathbf{D}_{ii} = d_{i}$. The \emph{unnormalised graph Laplacian} matrix is defined as
\begin{equation}
    \mathbf{L} = \mathbf{D} - \mathbf{W}.
\end{equation}
As $G$ is undirected, $w_{ij}=w_{ji}$, which means that the weight matrix is symmetric and hence $\mathbf{W}$ is symmetric and given $\mathbf{D}$ is symmetric, the unnormalized Laplacian is also symmetric. As symmetric matrices are special cases of normal matrices, they are Hermitian matrices and have real eigenvalues. Another common characterisation of the Laplacian matrix is the \emph{normalised Laplacian} \citep{chung1997spectral},
\begin{equation} \label{lnorm}
    \mathbf{L}_{\mathrm{norm}} = \mathbf{D}^{-\frac{1}{2}}\mathbf{L}\mathbf{D}^{-\frac{1}{2}} = \mathbf{I} - \mathbf{W}_{\mathrm{norm}} = \mathbf{I} - \mathbf{D}^{-\frac{1}{2}}\mathbf{W}\mathbf{D}^{-\frac{1}{2}},
\end{equation}
where $\mathbf{W}_{\mathrm{norm}}$ is known as the normalised adjacency matrix \footnote{Strictly speaking, the second equality only holds for graphs without isolated vertices.}. The spectrum of the graph is defined as the density of the eigenvalues of the given adjacency, Laplacian or normalised Laplacian matrices corresponding to the graph. Unless otherwise specified, we will consider the spectrum of the normalised Laplacian.


\section{Motivations for A New Approach on Comparing the Spectra of Large Graphs}\label{subsec:motivations}

In this section, without loss of generality, we motivate the need for a better approach for spectral density approximation in the context of the problem of comparing large graphs.

\subsection{Graph comparison using iterative algorithms}

For large sparse graphs, with millions or billions of nodes, learning the exact spectrum using eigen-decomposition is unfeasible due to the $\mathcal{O}(n^{3})$ cost.
Powerful iterative methods, such as the Lanczos algorithm, are then proposed as cost-efficient alternatives to approximate the graph spectrum with a sum of weighted Dirac delta functions closely matching the first $m$ moments (explained in Appendix \ref{subsec: lanczos}) of the spectral density \cite{Ubaru2016}:
\begin{equation}
    p(\lambda) = \frac{1}{n}\sum_{i=1}^{n}\delta(\lambda-\lambda_{i}) \approx \sum_{i=1}^{m}w_{i}\delta(\lambda-\lambda_{i}),
\end{equation}
where $\sum_{i=1}^m w_{i} = 1$, and $\lambda_{i}$ denotes the $i$-th eigenvalue in the spectrum.

However, such an approximation is undesirable because natural divergence measures between densities, such as the relative entropy $\mathcal{D}_{\mathrm{KL}}(p||q) \in (0,\infty)$ from the fields of information theory \cite{cover2012elements} and information geometry \cite{amari2007methods} shown in equation \eqref{eq:kldivergence}
\begin{equation}
\label{eq:kldivergence}
    \mathcal{D}_{\mathrm{KL}}(p||q) =
    \int p(\lambda)\log \frac{p(\lambda)}{q(\lambda)} d\lambda
\end{equation}
is infinite for densities that are mutually singular. The use of the Jensen-Shannon divergence simply re-scales the divergence into $\mathcal{D}_{\mathrm{JS}}(p||q) \in (0,1)$. This puts us in the counter-intuitive scenarios, such as,
\begin{itemize}
    \item an infinite (or maximal) divergence upon the removal or addition of a single edge or node in a large network;
    \item an infinite (or maximal) divergence between two graphs generated using the same random graph model and identical hyper-parameters.
\end{itemize}
This does not comply with our notion of network similarity. Two networks generated from the same stochastic process with the same hyper-parameters are highly similar and hence should have a low divergence. Similarly, the removal of an edge in a large network, such as for example two people un-friending each other on a large social network, would not in general be considered a fundamental change in the network structure.

One way to circumvent the above problem is to use kernel smoothing. However, as we argue in the following, this results in losing the original moment information.

\subsection{On the Importance of Moments}  \label{subsec:momentsmatter}

Given that all iterative methods essentially generate a $m$ moment empirical spectral density (ESD) approximation, it is instructive to ask what information is contained within the first $m$ spectral moments. 

To answer this question concretely, we consider the spectra of random graphs. By investigating the finite size corrections and convergence of individual moments of the empirical spectral density (ESD) compared to those of the limiting spectral density (LSD), we see that the observed spectra are faithful to those of the underlying stochastic process. Put simply, given a random graph model, if we compare the moments of the spectral density observed from a single instance of the model to that averaged over many instances, we see that the moments we observe are informative about the underlying stochastic process.

\subsubsection{ESD moments converge to those of the LSD}

For random graphs, with independent edge creation probabilities, their spectra can be studied through the machinery of random matrix theory \cite{akemann2011oxford}.

We consider the entries of an $n\times n$ matrix $\mathbf{X}_n$ to be zero mean and independent, with bounded moments. For such a matrix, a natural scaling which ensures we have bounded norm as $n \rightarrow \infty$ is $\mathbf{X}_{n} = \mathbf{M}_{n}/\sqrt{n}$.  
It can be shown (see for instance \cite{feier2012methods}) that the moments of a particular instance of a random graph and the related random matrix $\mathbf{X}_n$ converge to those of the limiting counterpart in probability with a correction of $\mathcal{O}(n^{-2})$.

\subsubsection{Finite size corrections to moments get worse with larger moments}

 A key result, akin to the normal distribution for classical densities, is the semi-circle law for random matrix spectra \cite{feier2012methods}. For matrices with independent entries $a_{ij}$, $\forall i>j$, with common element-wise bound $K$, common expectation $\mu$ and variance $\sigma^{2}$, and diagonal expectation $\mathbb{E}a_{ii}=\nu$, it can be shown that the corrections to the semi-circle law for the moments of the eigenvalue distribution,
\begin{equation}
 \int x^{m}d\mu(x) = \frac{1}{n}\text{Tr}\mathbf{X}^{m}_{n},
\end{equation}
have a corrective factor bounded by \cite{furedi1981eigenvalues}
\begin{equation}
    \frac{K^{2}m^{6}}{2\sigma^{2}n^{2}}.
\end{equation}
Hence, the finite size effects are larger for higher moments than that for the lower counterparts. This is an interesting result, as it means that for large graphs with $n \rightarrow \infty$, the lowest order moments, which are those learned by any iterative process, best approximate those of the underlying stochastic process.

\subsection{The argument against kernel smoothing}\label{subsec:kernelsmoothingbad}

To alleviate the limitations explained in this section, practitioners typically generate a smoothed spectral density by convolving the Dirac mixture with a smooth kernel \cite{takahashi2012discriminating,banerjee2008spectrum}, typically Gaussian or Cauchy,
\begin{eqnarray} 
    \tilde{p}(\lambda) &=& \int k_\sigma(\lambda-\lambda') p(\lambda') d\lambda'  \nonumber \\ 
    &=&
    \int k_\sigma(\lambda-\lambda')\sum_{i=1}^{n}w_{i}\delta(\lambda'-\lambda_{i})d\lambda'  \nonumber \\
    &=& \sum_{i=1}^{n}w_{i}k_\sigma(\lambda-\lambda_{i}), \label{eq:smootheddensity}
\end{eqnarray}
or simply a histogram \cite{banerjee2008spectrum} to facilitate visualisation and comparison. However, this introduces hyperparameters, such as the choice of convolving kernel, the smoothing parameter or the number of bins for the histogram, which heavily affect the resolution of the spectra. To show this, the moments of the Dirac mixture are given as, 
\begin{equation}
    \langle \lambda^{m} \rangle = \sum_{i=1}^{n}w_{i}\int \delta(\lambda-\lambda_{i})\lambda^{m}d\lambda = \sum_{i=1}^{n}w_{i}\lambda_{i}^{m}.
\end{equation}
For simplicity we assume that the kernel function is symmetric, defined on the real line, and permits all moments, which are satisfied for the commonly used Gaussian kernel. Consider the moments of the modified smooth function in \eqref{eq:smootheddensity}
\begin{equation}
    \begin{aligned}
    \langle \tilde{\lambda}^{m} \rangle & =\sum_{i=1}^nw_{i}\int k_\sigma(\lambda-\lambda_{i})\lambda^{m}d\lambda \\
    & = \sum_{i=1}^nw_{i}\int k_\sigma(\lambda')(\lambda'+\lambda_{i})^{m}d\lambda'\\
    & = \langle \lambda^{m} \rangle + \sum_{i=1}^{n}w_{i} \sum_{j=1}^{r/2} {r \choose 2j}\mathbb{E}_{k_\sigma(\lambda)}(\lambda^{2j})\lambda_{i}^{m-2j},
    \end{aligned}
\end{equation}
where the sum over $j$ is up to $r/2$, with $r$ being $m$ if $m$ is even and $m-1$ if $m$ is odd. Further, $\mathbb{E}_{k_\sigma(\lambda)}(\lambda^{2j})$ denotes the $2j$-th central moment of the kernel function $k_\sigma(\lambda)$. We have used the fact that the infinite domain is invariant under shift re-paramatrisation and that the odd moments of any symmetric distribution are $0$. This proves that kernel smoothing alters moment information and that this process gets more pronounced for higher moments. Furthermore, given that $w_{i} > 0$, $\mathbb{E}_{k_\sigma(\lambda)}(\lambda^{2j}) > 0$ and, for the normalised Laplacian with $\lambda_{i} > 0$, the corrective term is manifestly positive and so the smoothed moment estimates are biased. For large random graphs, the moments of a generated instance converge to those averaged over many instances \cite{feier2012methods}, hence by biasing our moment information we limit our ability to learn about the underlying stochastic process.


\section{The Method of Maximum Entropy}\label{subsec: maxent_method}

The method of maximum entropy, hereafter referred to as \emph{MaxEnt} \cite{Presse2013}, is a procedure for generating the most conservative density estimate, with respect to the uniform distribution. It can be seen as the maximally uncertain probability distribution possible with the given information. 
To determine the spectral density $p(\lambda)$ using MaxEnt, we maximise the entropic functional
\begin{equation} \label{BSG}
    S = - \int p(\lambda)\log p(\lambda)d\lambda- \sum_{i}\alpha_{i}\bigg[\int p(\lambda)\lambda^{i}d\lambda - \mu_{i}\bigg],
\end{equation}
with respect to $p(\lambda)$, where $ \mathbb{E}_{p}[ \lambda^{i}]= \mu_{i}$ are the power moment constraints on the spectral density, which are estimated using stochastic trace estimation as explained in section \ref{subset: stoctrace}. Algorithm \ref{alg:preprocessing} presents the procedure to learn them. The resultant MaxEnt spectral density has the form
\begin{equation}
    p(\lambda \vert \{ \alpha_{i} \}) = \exp[-(1+\sum_{i}\alpha_{i} \lambda^{i})],
\end{equation}
where the coefficients $\{ \alpha_{i} \}_{i=1}^{m}$ are derived from optimising \eqref{BSG}. For simplicity, we denote $p(\lambda \vert \{ \alpha_{i}\}_{i=1}^{m})$ as  $p(\lambda )$. We develop a novel algorithm (Algorithm \ref{alg:maxent}), where NCG denotes Newton conjugate gradient, for determining the density of maximum entropy. We use analytical expressions for the gradient and Hessian explicitly as opposed to approximately \cite{bandyopadhyay2005maximum}. We note that the method of maximum entropy has been used previously for calculating the spectra of large sparse Hamiltonians in Physics, and has been shown to be far superior in terms of moment information than kernel polynomial methods \cite{silver1997calculation}. Therefore, we do not compare our method against kernel polynomial methods in this paper.

\subsection{Analytic forms for the differential entropy and divergence from MaxEnt}\label{subsec: analytic_diff_ent_and_divergence}

To calculate the differential entropy we simply note that
\begin{equation}
    \mathcal{S}(p) = \int p(\lambda) (1+\sum_{i}^{m}\alpha_{i}\lambda^{i})d\lambda = 1+\sum_{i}^{m}\alpha_{i}\mu_{i}.
\end{equation}
\noindent The KL divergence between two MaxEnt spectra, $p(\lambda) = \exp[-(1+\sum_{i}\alpha_{i}\lambda^{i})]$ and $q(\lambda) = \exp[-(1+\sum_{i}\beta_{i}\lambda^{i})]$, can be written as,
\begin{equation}
    \mathcal{D}_{\mathrm{KL}}(p||q) = \int p(\lambda)\log \frac{p(\lambda)}{q(\lambda)}d\lambda = -\sum_{i}(\alpha_{i}-\beta_{i})\mu_{i}^{p},
\end{equation}
where $\mu_{i}^{p}$ refers to the $i$-th moment constraint of the density $p(\lambda)$. Similarly, the symmetric-KL divergence can be written as,
\begin{equation}
    \frac{\mathcal{D}_{\mathrm{KL}}(p||q)+\mathcal{D}_{\mathrm{KL}}(q||p)}{2} = \frac{\sum_{i}(\alpha_{i}-\beta_{i})(\mu_{i}^{q}-\mu_{i}^{p})}{2},
\end{equation}
where all the $\alpha$ and $\beta$ are derived from the optimisation and all the $\mu$ are given from the stochastic trace estimation.

\subsection{Stochastic trace estimation} \label{subset: stoctrace}

The intuition behind stochastic trace estimation is that we can accurately approximate the moments of $\lambda$ with respect to the spectral density $p(\lambda)$ by using computationally cheap matrix-vector multiplications. The moments of $\lambda$ can be estimated as,
\begin{equation}
    n\,\mathbb{E}_{p}(\lambda^{m}) =\mathbb{E}_{\mvec{v}}(\mvec{v}^{T} \mvec{X}^{m}\mvec{v}) \approx \frac{1}{d}\sum_{j=1}^{d}\mvec{v}_{j}^{T}\mvec{X}^{m}\mvec{v}_{j},
\end{equation}
where $\mvec{v}_j$ is any random vector with zero mean and unit covariance and $\mvec{X}$ is a $n \times n$ matrix whose eigenvalues are $\{ \lambda_i \}_{i=1}^n$. This enables us to efficiently estimate the moments in $\mathcal{O}(d\times m\times n_\mathrm{nz})$ for sparse matrices, where $d\times m \ll n$. We use these as moment constraints in our MaxEnt formalism to derive the functional form of the spectral density. Examples of this in the literature include \cite{ubaru2017fast,ete}.

\begin{algorithm}[htb!]
\caption{Learning the Graph Laplacian Moments}
\label{alg:preprocessing}
\begin{algorithmic}[1]
		\STATE {\bfseries Input:} Normalized Laplacian $\mvec{L}_{\mathrm{norm}}$, Number of Probe Vectors $d$, Number of moments required $m$
		\STATE {\bfseries Output:} Moments of Normalised Laplacian $\{\mu_{i}\}$
		\FOR {$i$ in $1,\dots,d$}
		\STATE  Initialise random vector $\mvec{z}_{i}\in R^{1\times n}$
		\FOR {$j$ in $1,\dots,m$}
		\STATE $\mvec{z}_{i}' = \mvec{L}_{\mathrm{norm}}\mvec{z}_{i}$
		\STATE $\rho_{ij} =  \mvec{z}_{i}\Trans \mvec{z}_{j}'$
		\ENDFOR
		\ENDFOR
		\STATE $\mu_{i} = 1/d \times \sum_{j=1}^{d}\rho_{ij}$ 
\end{algorithmic}
\end{algorithm}

\subsection{The Entropic Spectral Learning algorithm}

Algorithm \ref{alg:preprocessing} learns the moments of the normalised graph Laplacian. The appropriate Lagrange multipliers for the maximum entropy density are learned with algorithm \ref{alg:maxent}.
The full algorithm, which takes the matrix as an input and gives the maximum entropy spectral density, is then summarized in Algorithm \ref{alg:ESL}.

\begin{algorithm}[htb!]
	\caption{MaxEnt Algorithm}
	\label{alg:maxent}
	\begin{algorithmic}[1]
		\STATE {\bfseries Input:} Moments $\{\mu_{i}\}$, Tolerance $\epsilon$, Hessian noise $\eta$
		\STATE {\bfseries Output:} Coefficients $\{\alpha_{i}\}$
		\STATE Initialize $\alpha_{i} = 0$.
		\STATE Minimize $\mathcal{S}(p) = \int_{0}^{1}p_{\alpha}(\lambda)d\lambda + \sum_{i}\alpha_{i}\mu_{i}$
		\STATE Gradient $\mvec{g} = \nabla \mathcal{S}(p)$; \quad $g_j=\mu_{j}-\int_{0}^{1}p_{\alpha}(\lambda)\lambda^{j}d\lambda$
		\STATE Hessian $ \mvec{H} = (\tilde{\mvec{H}}+\tilde{\mvec{H}}^T)/2 + \eta; \quad \tilde{H}_{jk}= \int_{0}^{1}p_{\alpha}(\lambda)\lambda^{j+k}d\lambda$
		\WHILE {not $\forall j \, g_{j} < \epsilon$}
		    \STATE NCG($\mathcal{S},\vec{g},\mvec{H}$)
		\ENDWHILE
	\end{algorithmic}
\end{algorithm}

\begin{algorithm}[htb!]
	\caption{Entropic Spectral Learning (ESL)}
	\label{alg:ESL}
	\begin{algorithmic}[1]
		\STATE {\bfseries Input:} Normalized Laplacian $\mvec{L}_{\mathrm{norm}}$, Number of Probe Vectors $d$, Number of moments required $m$, Tolerance $\epsilon$, Hessian noise $\eta$
		\STATE {\bfseries Output:} MaxEnt Spectral Density (MESD) $ p(\lambda)$
		\STATE Moments $\{ \mu_i \}_{i=1}^m$ $\leftarrow$ Algorithm \ref{alg:preprocessing} $\left(\mvec{L}_{\mathrm{norm}}, d, m \right)$
		\STATE MaxEnt coefficients $\{ \alpha_i \}_{i=1}^m$ $\leftarrow$ Algorithm \ref{alg:maxent} $\left(\{ \mu_i \}_{i=1}^m, \epsilon, \eta \right)$
		\STATE MaxEnt Spectral Density $p(\lambda) = \exp[-(1+\sum_{i}\alpha_{i}\lambda^{i})]$
	\end{algorithmic}
\end{algorithm}


\section{Visualising the modelling power of MaxEnt}\label{sec:modelling_power_of_maxent}

Having developed a theory for why a smooth exact moment matched approximation of the spectral density is crucial to learning the characteristics of the underlying stochastic process, and having proposed a set of Algorithms \ref{alg:preprocessing} and \ref{alg:maxent} to learn such a density, we test the practical utility of our method and algorithm on examples where the limiting spectral density is known.

\subsection{The semi-circle law} \label{subsec:semicircle_and_beyond}

For Erd\H{o}s-R\'{e}nyi graphs with edge creation probability $p \in (0,1)$, and $np \rightarrow \infty$, the limiting spectral density of the normalised Laplacian converges to the semi-circle law and its Laplacian converges to the free convolution of the semi-circle law and $\mathcal{N}(0,1)$ \cite{jiang2012empirical}.

\begin{figure}[tbp]
	\centering
    \includegraphics[trim=1.0cm 0.1cm 1.0cm 0.5cm, clip, width=1.0\linewidth]{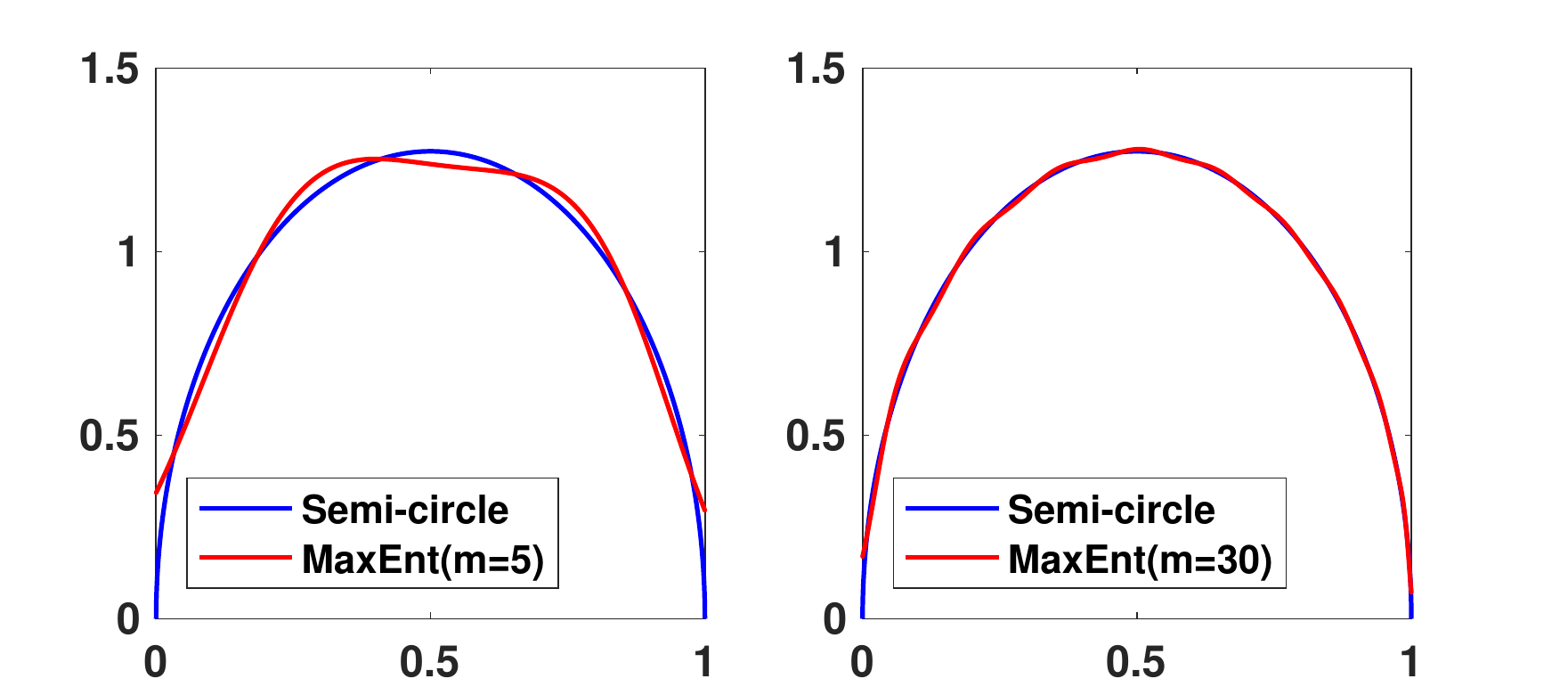}
	\caption{Maximum Entropy distribution fit to semi-circle density that is centered at 0.5 and has a radius of 0.5 $[x_{0},R] = [0.5,0.5]$ for different moment number $m$.}
	\label{fig:semicirclepure}
\end{figure}

\begin{figure}[tbp]
	\centering
	\includegraphics[trim=0.1cm 0cm 1.0cm 0cm, clip, width=1\linewidth]{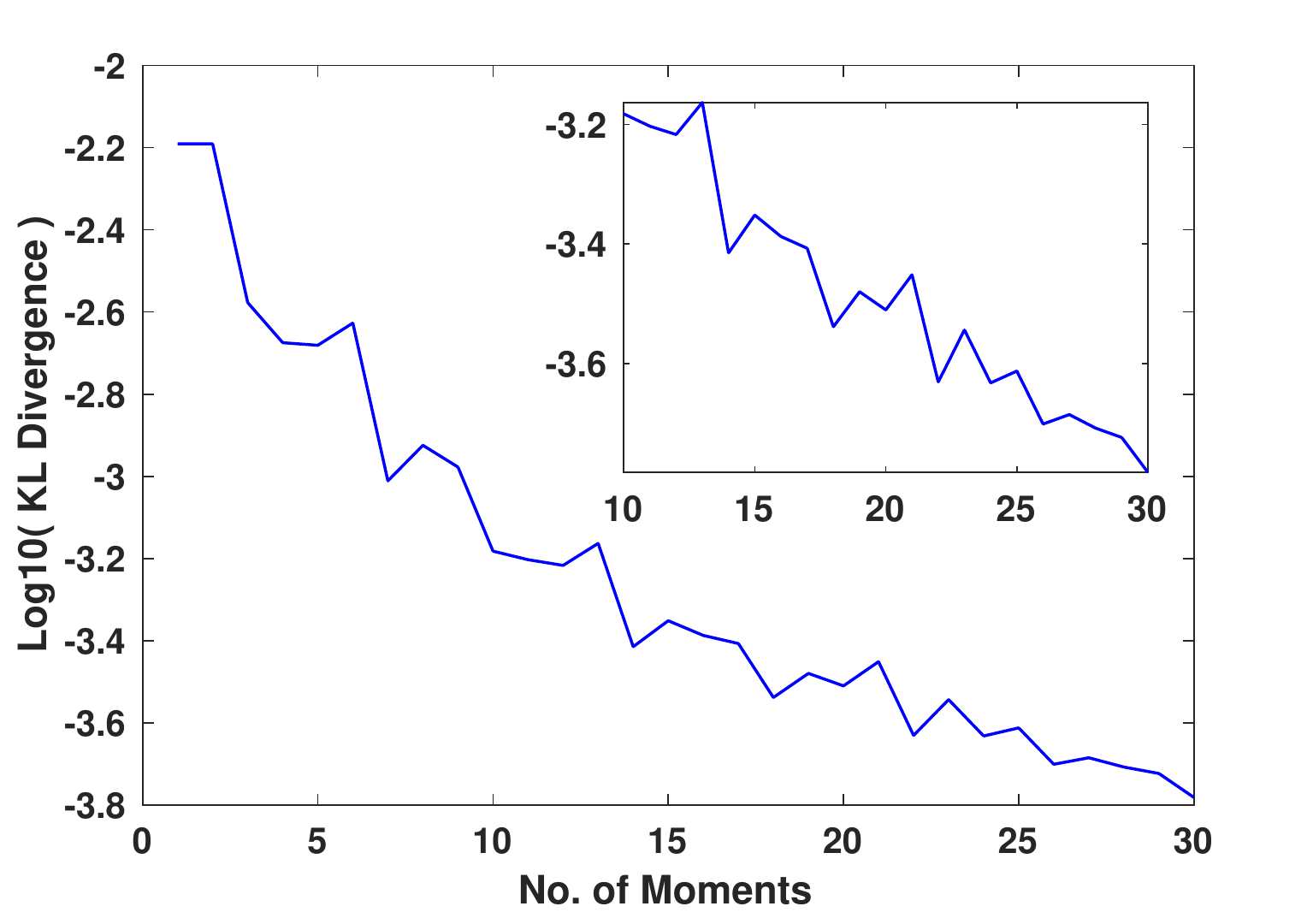}
	\caption{KL divergence between the Maximum Entropy distribution fit and the semi-circle density with zoom in to log scale for $10 \leq m \leq30$}\label{fig:maxentsemicirclekl}
\end{figure}

We consider here to what extent a MaxEnt distribution with finite moments can effectively approximate the density. Wigner's density is fully defined by its infinite number of central moments given by $\mathbb{E}_{\mu}(\lambda^{2n}) = (R/2)^{2n}C_{n}$, where $C_{n}\times (n+1) = {2n \choose n}$ are known as the Catalan numbers. As a toy example we generate a semi circle centered at $\lambda=0.5$ with $R=0.5$ and give the moments analytically to the maximum entropy algorithm. As can be seen in FIG \ref{fig:semicirclepure}, for $m=5$ moments, the central portion of the density is already well approximated, but the end points are not. This is largely corrected for $m=30$ moments. 
We plot the relative entropy between the semi-circle density and its maximum entropy surrogate, which we show in FIG \ref{fig:maxentsemicirclekl}. This shows that even a very modest number of moments can give an excellent fit. 

\subsection{Application to Erd\H{o}s-R\'{e}nyi normalised Laplacian} \label{subsec:semicircle_exp}

\begin{figure}[tbp]	
	\centering
	 \includegraphics[trim=1.3cm 0.7cm 1.5cm 0cm, clip, width=1.0\linewidth]{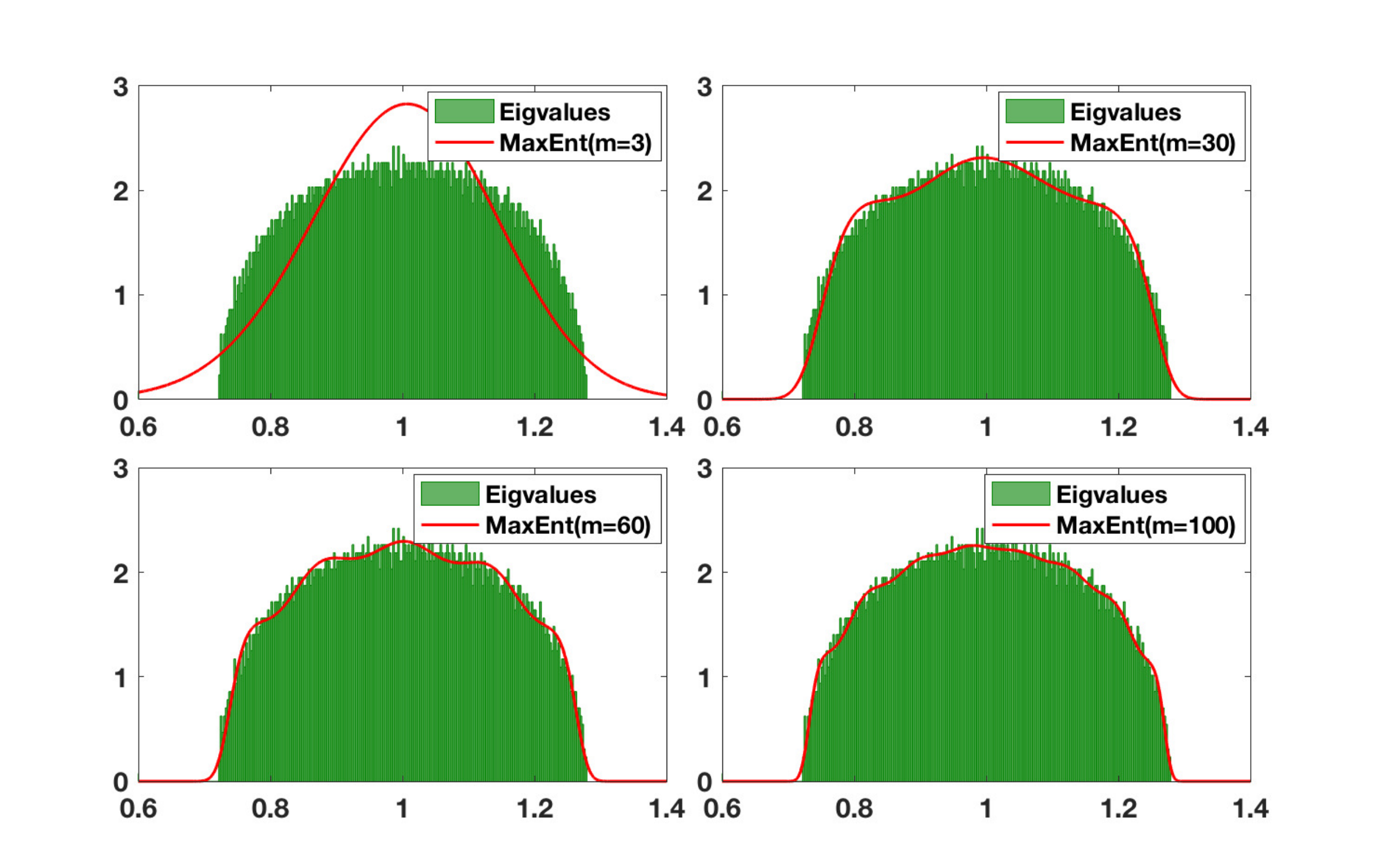}
	\caption{Maximum Entropy distribution fit to randomly generated $p=0.001, n = 5000$ Erd\H{o}s-R\'{e}nyi graph. The number of moments used for computing Maximum Entropy distributions increases from $m=3$ to $m=100$ abd the number of bins used for the eigenvalue histogram is $n_{b}=500$.}\label{fig:maxenter}
\end{figure}

We generate a Erd\H{o}s-R\'{e}nyi graph with $p=0.001$ and $n=5000$, and learn the moments using stochastic trace estimation. We then compare the fit between the MaxEnt distribution computed using a different numbers of input moments $m = 3,35,100$ and the graph eigenvalue histogram computed by eigen-decomposition. We plot the results in FIG \ref{fig:maxenter}. One striking difference between this experiment and the previous one is the number of moments needed to give a good fit. This can be seen especially clearly in the top left subplot of FIG \ref{fig:maxenter}, where the 3 moment, i.e Gaussian approximation, completely fails to capture the bounded support of the spectral density. Given that the exponential polynomial density is positive everywhere, it needs more moment information to learn the regions of boundedness of the spectral density in its domain. In the previous example we artificially alleviated this phenomenon by putting the support of the semi-circle within the entire domain. It can be clearly seen that increasing moment information successively improves the fit to the support FIG \ref{fig:maxenter}. Furthermore, the magnitude of the oscillations, which are characteristic of an exponential polynomial function, decay in magnitude for larger moments. 

\subsection{Beyond the semi-circle law} \label{subsec:ba_networks}

\begin{figure}[t]	
	\centering
	\includegraphics[trim=1.3cm 0.0cm 1.7cm 0.5cm, clip, width=1.0\linewidth]{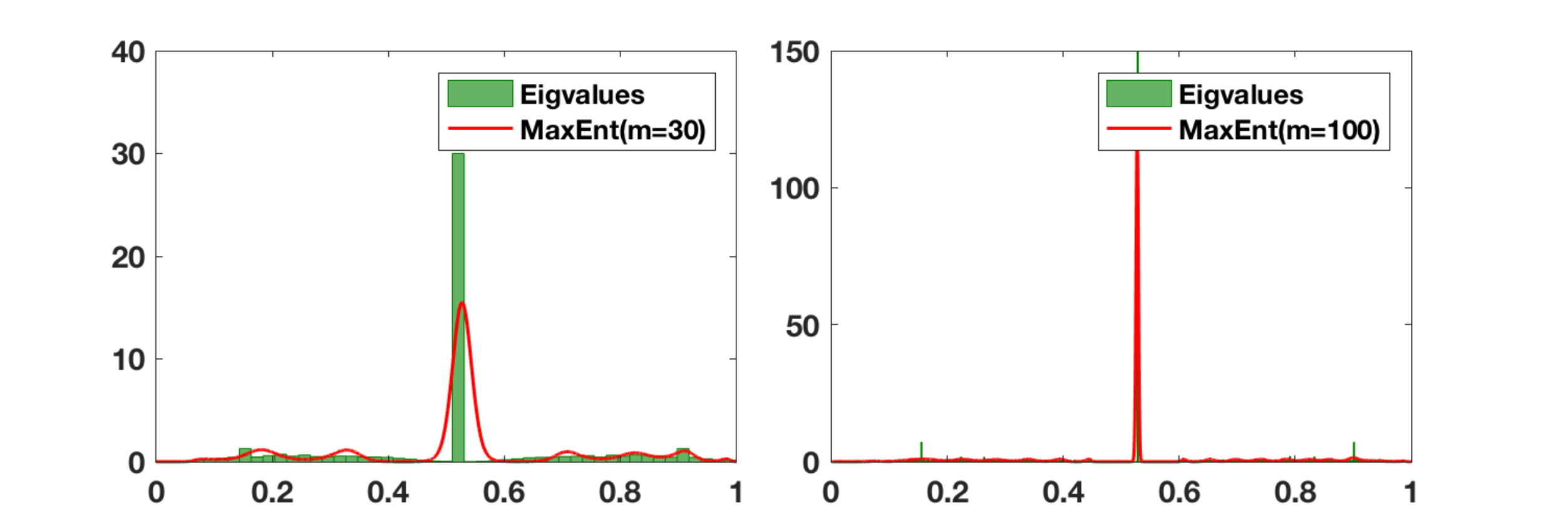}
	\caption{Maximum Entropy distribution fit to randomly generated $n = 5000$ Barab\'{a}si-Albert graph. The number of moments used for computing Maximum Entropy distributions and the number of bins used for the eigenvalue histogram are $m=30$, $n_{b}=50$ (Left) and $m=100$, $n_{b}=500$ (Right).}\label{fig:maxentba}
\end{figure}

For the adjacency matrix of an Erd\H{o}s-R\'{e}nyi graph with $p \propto 1/n$, the limiting spectral density does not converge to the semi-circle law and has an elevated central portion, and the scale free limiting density converges to a triangle like distribution \cite{farkas2001spectra}. For other random graph, such as the Barab\'{a}si-Albert \cite{barabasi1999emergence} also known as the scale free network, the probability of a new node being connected to a certain existing node is proportional to the number of links that existing node already has, violating the independence assumption required to derive the semi-circle density. We plot a Barab\'{a}si-Albert network ($n=5000$) and, similar to section \ref{subsec:semicircle_exp}, we learn the moments using stochastic trace estimation and plot the resulting MaxEnt spectral density against the eigenvalue histogram, as shown in FIG \ref{fig:maxentba}. For the Barab\'{a}si-Albert network, due to the extremity of the central peak, a much larger number of moments is required to get a reasonable fit. We also note that increasing the number of moments is essentially akin to increasing the number of bins in terms of spectral resolution, as seen in FIG \ref{fig:maxentba}.

\section{MaxEnt for computing graph similarity}\label{sec:all_experimetns}

In this section, we test the performance of our MaxEnt-based method for computing similarity between different types of synthetic and real world graphs. We first investigate the feasibility in recovering the parameters of random graph models, and then move onto classifying the network type as well as computing graph similarity by measuring the symmetric KL divergence among various synthetic and real world graphs.

\subsection{Inferring parameters of random graph models}

We investigate whether one can recover the network parameter values of a graph via its MaxEnt spectral density. We generate a random graph of a given size and parameter value (e.g., $n=50, p=0.6$) and learn its MaxEnt spectral characterisation. Then, we generate another graph of the same size but learn its parameter value by minimising the symmetric-KL divergence between its MaxEnt spectral surrogate and that of the original graph. We repeat the above procedures for different random graph models and different graph sizes ($n=50, 100, 150$), and the results are shown in Table \ref{table:learn_synnet_para}. It can be seen that  given simply the approximate MaxEnt spectrum, we are able to learn rather well the parameters of the graph producing that spectrum. Determining which random graph models best fit real world networks, characterised by their spectral divergence, so as to better understand their dynamics and characteristics has been explored in biology \citep{takahashi2012discriminating}, a potential application domain for our method.

\begin{table}[htb!]
	\caption{Average parameters estimated by MaxEnt for the 3 types of network. $n$ denotes the number of nodes in the network. $n$ denotes the number of nodes in the network.}\label{table:learn_synnet_para}
	\begin{center}
		\begin{small}
			\begin{sc}
				\begin{tabular}{lcccr}
					\toprule
					$n$ & 50  & 100  & 150 \\
					\midrule
					Erd\H{o}s-R\'{e}nyi ($p=0.6$)     & $0.600$   & $0.598$  & $ 0.604$   \\
					Watts-Strogatz ($p=0.4$)  & $0.468$   & $0.454$   & $0.414$   \\
					Barab\'{a}si-Albert ($r=0.4n$)   & $18.936$ & $40.239$  & $58.428$   \\
					\bottomrule
				\end{tabular}
			\end{sc}
		\end{small}
	\end{center}
	\vskip -0.1in
\end{table}

\subsection{Learning real world network types using MaxEnt and the symmetric KL divergence}\label{subsec: network_classification_maxent_kl}


As a real-world use case, we investigate which random network between Erd\H{o}s-R\'{e}nyi, Watts-Strogatz and Barab\'{a}si-Albert can best model the YouTube network from the SNAP dataset \cite{snapnets}. To this end, we compute the divergence between the MaxEnt spectral density of the YouTube network and those of the randomly generated graphs. We find, as shown in Table \ref{table:learn_real_para}, that the Barab\'{a}si-Albert gives the lowest divergence, which aligns with other findings for social networks \citep{barabasi1999emergence}.

\begin{table}[htb!]
	\caption{Minimum KL divergence between Entropic Spectrum of Youtube and that of synthetic networks}
	\label{table:learn_real_para}
	\begin{center}
		\begin{small}
			\begin{sc}
				\begin{tabular}{lccr}
					\toprule
					& Synthetic  & Youtube \\
					\midrule
					Erd\H{o}s-R\'{e}nyi  & $2.662$ &$7.728$\\
					Watts-Strogatz & $7.6123$ &$ 9.735$ \\
					Barab\'{a}si-Albert & $\mathbf{2.001}$ & $ \mathbf{7.593} $\\
					\midrule
				\end{tabular}
			\end{sc}
		\end{small}		
	\end{center}
	\vskip -0.1in
\end{table} 

\subsection{Comparing different real world networks}

\begin{figure}[t]
    \centering
    \includegraphics[width = 1.0\linewidth]{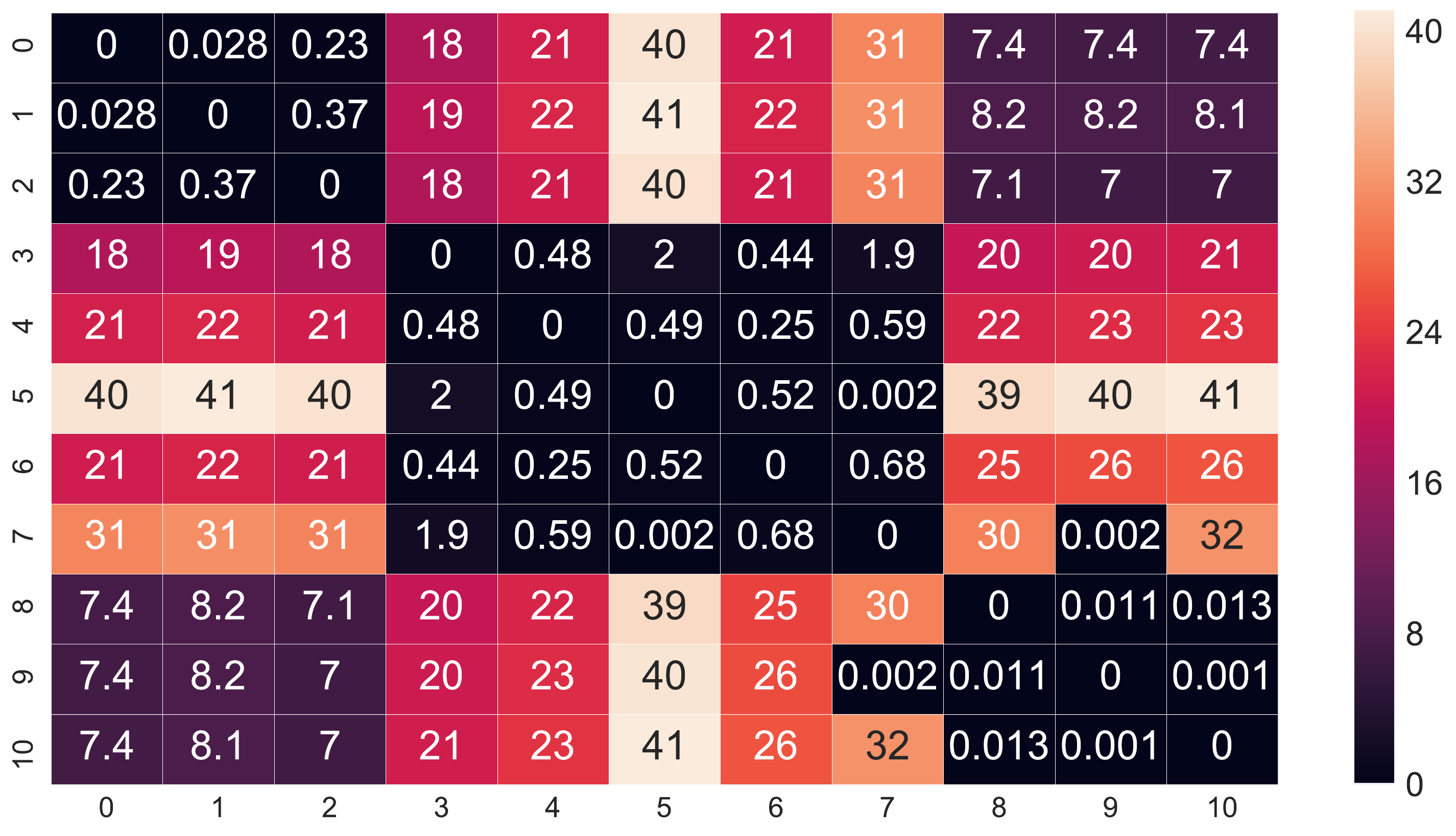}
    \caption{Symmetric KL heatmap between 9 graphs from the SNAP dataset, in ascending order [bio-human-gene1,
bio-human-gene2,
bio-mouse-gene,
ca-AstroPh,
ca-CondMat,
ca-GrQc,
ca-HepPh,
ca-HepTh,
roadNet-CA,
roadNet-PA,
roadNet-TX].}
    \label{fig:graphsinthewild}
\end{figure}

We now consider the feasibility of comparing real world graphs using their MESDs. Specifically, we take $3$ biological networks, $5$ citation networks and $3$ road networks from the SNAP dataset \cite{snapnets}, and compute the symmetric kl divergences between their MESDs with $m=100$ moments. We plot the results in the form of a heat map in FIG \ref{fig:graphsinthewild}. We see very clearly that the intra-class divergences between the biological, citation and road networks are much smaller than their inter-class divergences. This strongly suggests that the combination of a MESD and the symmetric KL divergence can be used to identify similarity in networks. Furthermore, as can be seen in the divergence between the human and mouse network, the spectra of the human genes are more closely aligned with each other than they are with the spectra of mouse genes. This suggests a reasonable amount of intra-class distinguishability as well. 

\section{MaxEnt for estimating cluster number}\label{subset:clustering_with_spec}

In this section, we discuss a second application of the proposed MaxEnt-based method, which is to estimate the number of clusters (i.e., communities) in a graph.
We first recall the following results from spectral graph theory \citep{chung1997spectral}.

\begin{proposition}
	\label{numberof0eigenvalues}
	Let G be an undirected graph with non-negative weights. Then the multiplicity $k$ of the eigenvalue $0$ of the Laplacian $L\in \mathcal{R}^{n\times n}$ is equal to the number of connected components $A_{1},...A_{k}$ in the graph. The eigenspace of the eigenvalue $0$ is spanned by the indicator vectors $\mathbbm{1}_{A_{1}},...,\mathbbm{1}_{A_{k}}$.
\end{proposition}

We refer the reader to \cite{von2007tutorial} for a simple proof of this result. Consequently, to learn the number of connected components in a graph, we simply count the number of $0$ eigenvalues. 
We now extend the idea behind this proposition, and consider groups of nodes containing far greater intra-group connections than inter-group connections as clusters. 
For small changes in the Laplacian, i.e., a very small number of links between the $k$ previously disconnected clusters, we expect from matrix perturbation theory \citep{bhatia2013matrix} the next $k-1$ smallest eigenvalues to be close to $0$. This is given by Weyl's bound on Hermitian matrices,
\begin{equation}
\begin{aligned}
    \Delta \lambda_{i}=|\lambda'_{i}-\lambda_{i}| \leq ||\mathbf{L}_{G'}-\mathbf{L}_G||,
\end{aligned}
\end{equation}
where $G'$ is the perturbed version of graph $G$ with a small number of edges between the previously disconnected $k$ clusters, $\lambda'_{i}$ are the perturbed eigenvalues, which differ from $\lambda_{i}$ (where $\lambda_{i}=0$ for $i=2,...,k$) by an amount $\Delta \lambda_{i}$, which is bounded by the norm of the difference matrix $\mathbf{L}_{G'}-\mathbf{L}_G$.
Notice that the bound holds for any consistent matrix norm. In the case of an entry-wise $L^1$-norm, for every edge added between the $i$-th and $j$-th node that are from previously separate clusters, the norm of the difference matrix goes as $2 \times |\sum_{g \in \mathcal{N}(i)} (\frac{1}{\sqrt{d_{g}d_{i}}}-\frac{1}{\sqrt{d_{g}(d_{i}+1)}}) + \sum_{h \in \mathcal{N}(j)} (\frac{1}{\sqrt{d_{h}d_{j}}}-\frac{1}{\sqrt{d_{h}(d_{j}+1)}}) + \frac{1}{\sqrt{(d_{i}+1)(d_{j}+1)}}| $, where $\mathcal{N}(i)$ and $\mathcal{N}(j)$ denote the neighborhood of $i$ and $j$ in $G$, respectively,
and $d_{i}$ denotes the degree of the $i$-th node.
Assuming that the degrees of the nodes in each cluster to be similar and $d_{i,j} \gg 1$, we see that for each added inter-cluster edge the bound grows as 
$ \mathcal{O} \big( d_{i}^{-1} + d_{j}^{-1} + (d_{i}d_{j})^{-\frac{1}{2}} \big) $.


\subsection{Motivations for a new approach on learning the number of clusters in large graphs}

For the case of large sparse graphs, where only iterative methods such as the Lanczos algorithm can be used, the same arguments of the previous section \ref{subsec:motivations} apply. 
This is because the delta functions are now weighted, and to obtain a reliable estimate of the eigengap, one must smooth the delta functions.

\subsection{Using MaxEnt to learn the number of clusters}\label{subsec:learn_no_clusters}

For our definition of a cluster described above, we would expect a smoothed spectral density plot to have a spike near $0$. We expect the moments of the spectral density to encode this information and the mass of this peak to be spread. We hence look for the first spectral minimum in the MaxEnt spectral density and calculate the number of clusters as shown in Algorithm \ref{alg:clusteralg}.

\begin{algorithm}[h]
	\caption{Algorithm for Cluster Number Estimation}
	\label{alg:clusteralg}
	\begin{algorithmic}[1]
		\STATE {\bfseries Input:} Lagrange Multipliers ${\alpha_{i}}$, Matrix Dimension $n$, Tolerance $\eta$
		\STATE {\bfseries Output:} Number of Clusters $N_{c}$
		\STATE Initialize $p(\lambda) = \exp{-[1+\sum_{i}\alpha_{i}x^{i}]}$.
		\STATE Minimize $\lambda*$ s.t $\frac{dp(\lambda)}{d\lambda}|_{\lambda=\lambda*} \leq \eta \thinspace \& \thinspace \frac{d^{2}p(\lambda)}{d\lambda^{2}} > 0$
		\STATE Calculate $N_{c} = n\int_{0}^{\lambda*} p(\lambda)d\lambda$	
	\end{algorithmic}
\end{algorithm}

\subsection{Experiments}\label{subsec: cluster_number_learning}

This set of experiments evaluates the effectiveness of our spectral method in Algorithm \ref{alg:clusteralg} for learning the number of distinct clusters in a network, where we compare it against the Lanczos algorithm with kernel smoothing on both synthetic and real-world networks.

\subsubsection{Synthetic networks}

The synthetic data consists of disconnected sub-graphs of varying sizes and cluster numbers, to which a small number of intra-cluster edges are added. We use an identical number of matrix vector multiplications, i.e., $m=80$ (see Appendix \ref{subsec:implementation_details} for experimental details for both MaxEnt and Lanczos methods), and estimate the number of clusters and report the fractional error. The results are shown in Table \ref{table:fractional_error_syn}. In each case, the method achieving lowest detection error is highlighted in bold. It is evident that the MaxEnt approach outperforms Lanczos as the number of clusters and the network size increase. 
We observe a general improvement in performance for larger graphs, visible in the differences between fractional errors for MaxEnt as the graph size increases and not kernel-smoothed Lanczos. 

\begin{table}[h]
	\caption{Fractional error in cluster number detection for synthetic networks using MaxEnt and Lanczos methods with 80 moments. $n_c$ denotes the number of clusters in the network and $n$ is the number of nodes.}\label{table:fractional_error_syn}
	\begin{center}
		\begin{small}
			\begin{sc}
				\begin{tabular}{lccr}
					\toprule
					$n_c$ ($n$) & Lanczos  & MaxEnt \\
					\midrule
					 9 (270)    & $ \mathbf{3.20 \times 10^{-3}}$   & $9.70 \times 10^{-3}$  \\
					30 (900) & $ 1.41\times 10^{-2}$   & $ \mathbf{6.40 \times 10^{-3}}$ \\
					90 (2700) & $1.81\times 10^{-2}$   & $\mathbf{5.80\times 10^{-3}}$   \\
					240 (7200) & $2.89\times 10^{-2}$   & $\mathbf{3.50\times 10^{-3}}$   \\
					\bottomrule
				\end{tabular}
			\end{sc}
		\end{small}
	\end{center}
	\vskip -0.1in
\end{table}

\begin{figure}[t]
	\centering
	\begin{subfigure}{0.5\linewidth}
		\centering
	    \includegraphics[trim=0cm 0cm 0.1cm 0.0cm, clip, width=1.0\linewidth]{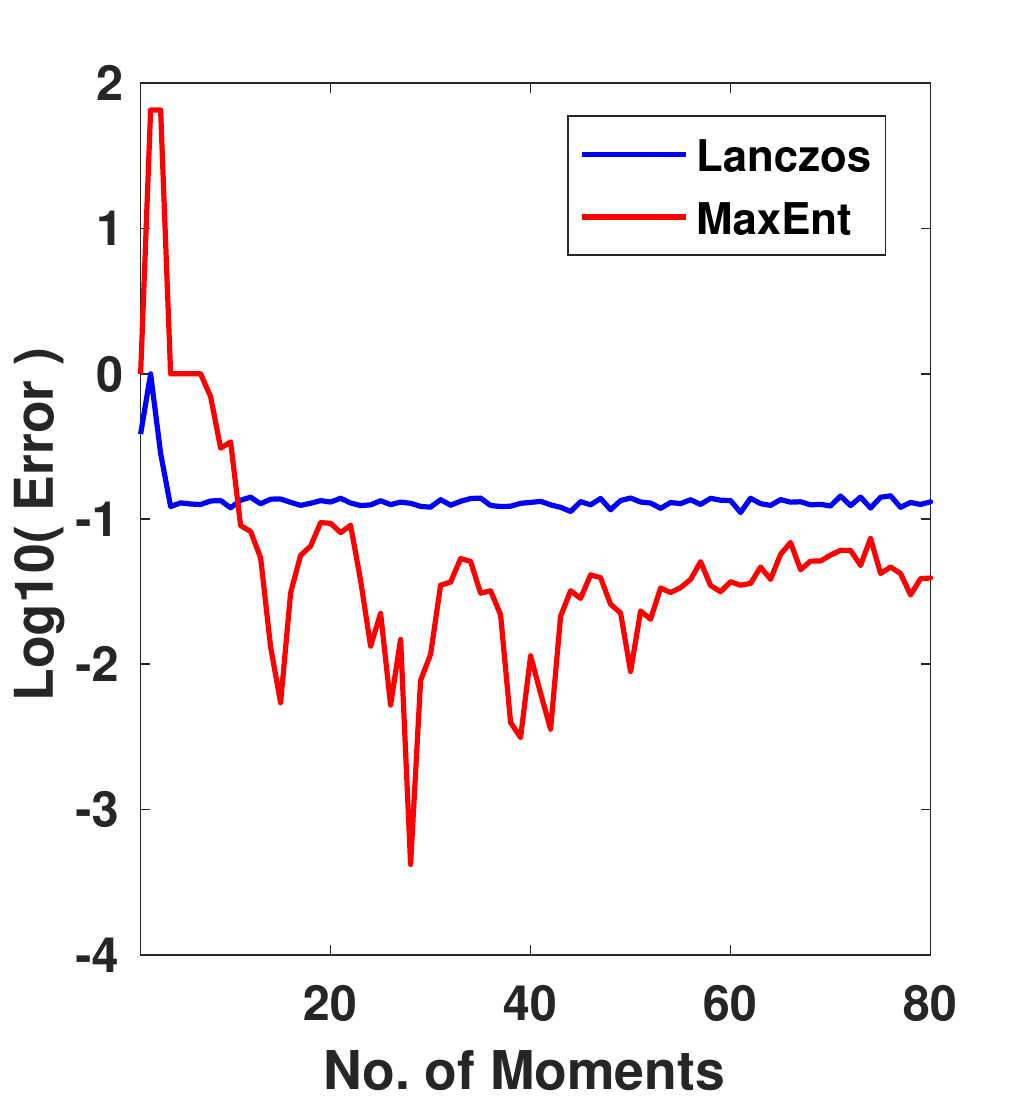}
	    \caption{305 clusters}
	    \label{subfig:emailerror1}	
	\end{subfigure}%
	\begin{subfigure}{0.5\linewidth}
		\centering
    	\includegraphics[trim=0cm 0cm 0.1cm 0.0cm, clip, width=1.0\linewidth]{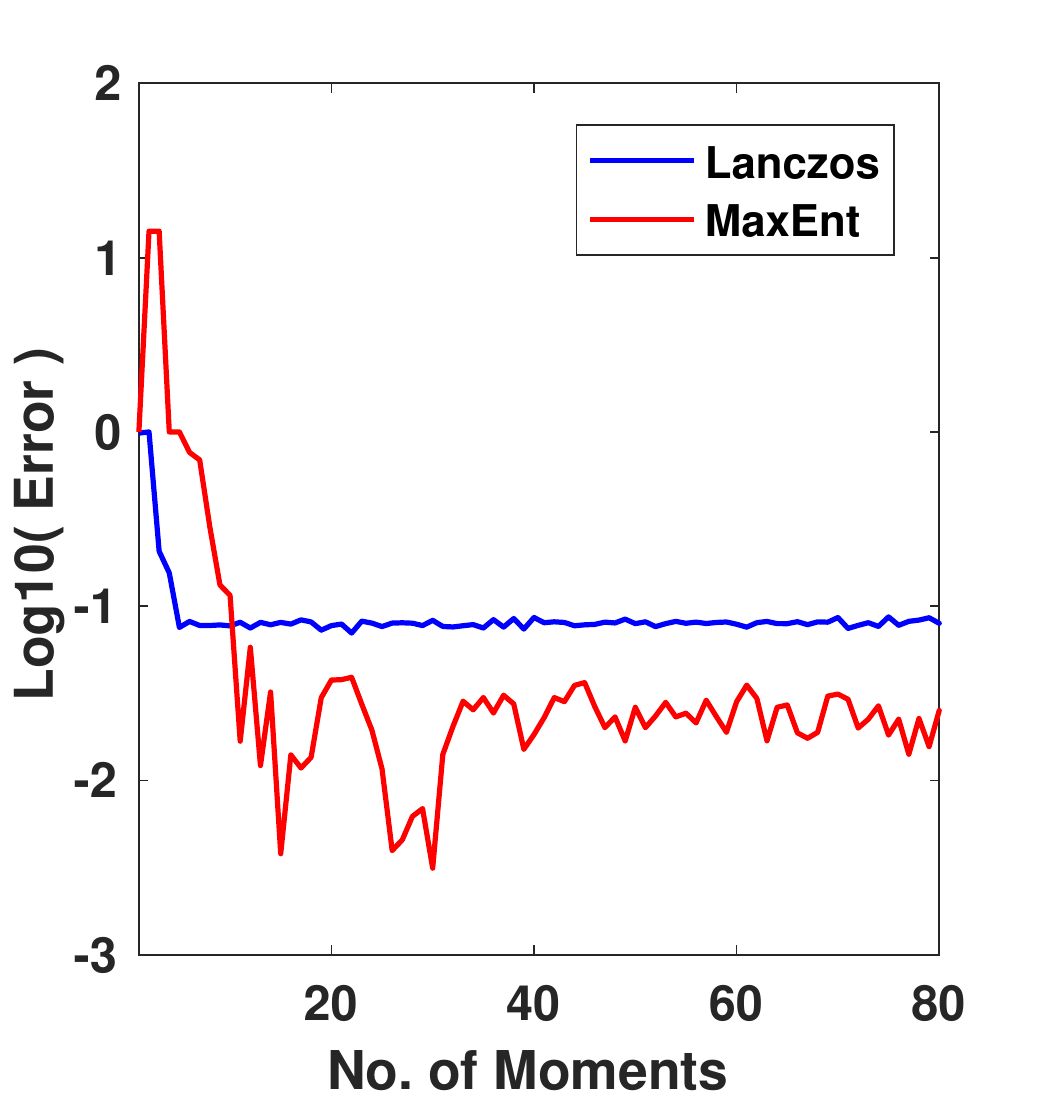}
	    \caption{1,355 clusters}
	    \label{subfig:emailerror2}	
	\end{subfigure}%
	\caption{Log error of cluster number detection using MaxEnt and Lanczos methods on large synthetic networks: a) synthetic network of  201,600 nodes and 305 clusters and b) synthetic network of 404,420 nodes and 1,355 clusters.}
	\label{fig:largeSyntheNet}	
\end{figure}

To test the performance of our approach for networks that are too large to apply eigen-decomposition, we generate two large networks by mixing the Erd{\"o}s-R{\'e}nyi, Watts-Strogatz and Barab\'{a}si-Albert random graph models. The first large network has a size of  201,600 nodes and comprises 305 interconnected clusters whose size varies from 500 to 1000 nodes. The second large network has a size of 404,420 nodes and comprises interconnected 1355 clusters whose size varies from 200 to 400 nodes. The results in FIG \ref{fig:largeSyntheNet} show that for both methods, the detection error generally decreases as more moments are used, and our maximum entropy approach again outperforms the Lanczos method for both large synthetic networks.

\subsubsection{Small real world networks}\label{subsubsec:small_real_network}

\begin{figure}[t]
	\centering
	\includegraphics[width=1\linewidth]{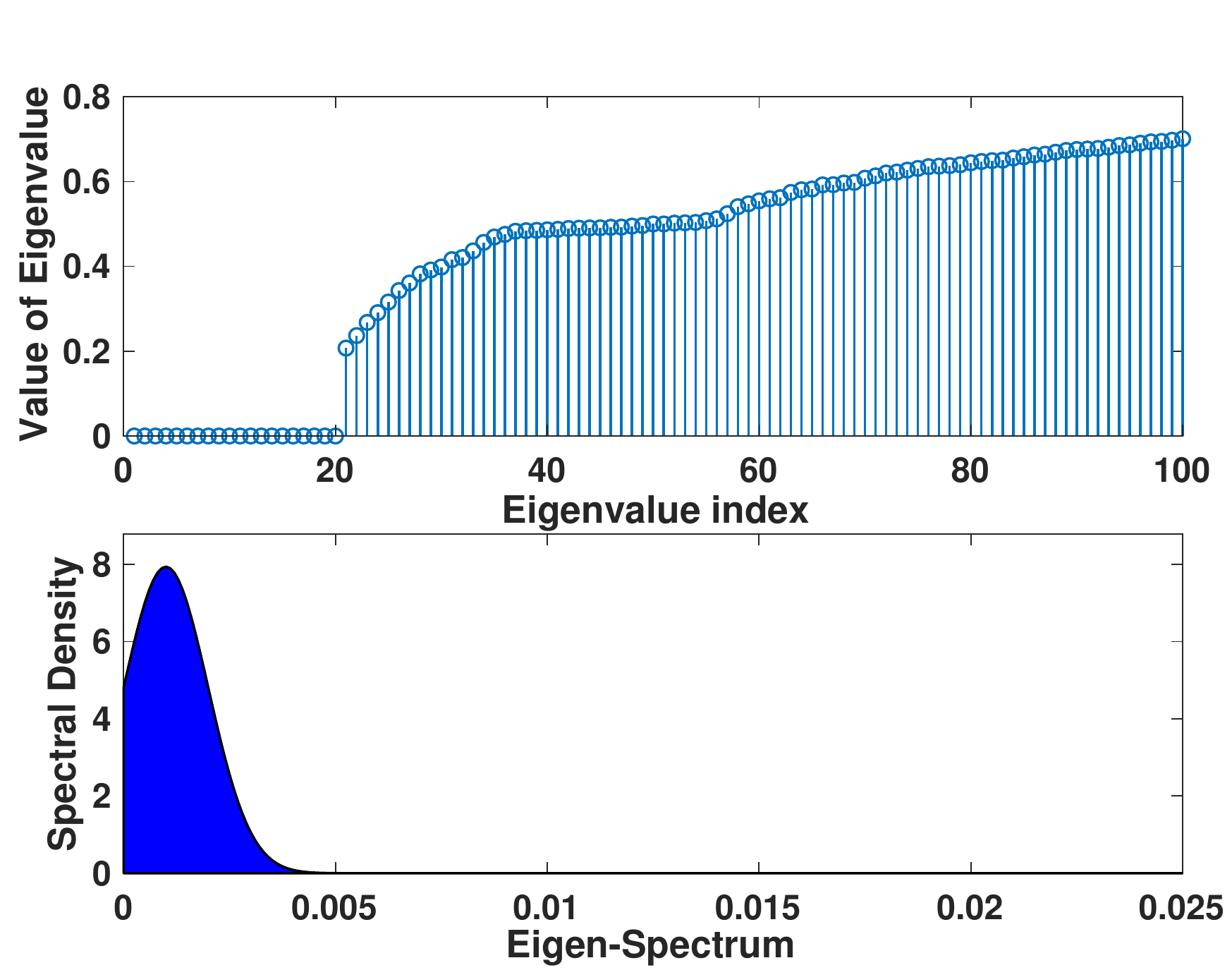}
	\caption{Eigenvalues of the Email dataset, with clear spectral gap along with the corresponding spectral density near the origin, showing a minimum at the value of the eigengap. The shaded area multiplied by the number of nodes $n$ predicts the number of clusters.}
	\label{fig:email}	
\end{figure}

We next experiment with relatively small real world networks, such as the Email network in the SNAP dataset, which is an undirected graph where the $n=1,003$ nodes represent members of a large European research institution and the edges represent the existence of email communication betweem them. For such network, we can still calculate the ground-truth number of clusters by computing the eigenvalues explicitly and finding the spectral gap near $0$. For the Email network, we count $20$ very small eigenvalues before a large jump in magnitude (measured in the log scale) and set this as the ground-truth. This is shown in FIG \ref{fig:email}, where we display the value of each of the eigenvalues in increasing order and how this results in a broadened peak in the MaxEnt spectrum. The area under the curve multiplied by the number of network nodes is the number of clusters $N_{c}$. 
We note that the number $20$ differs from the value of $42$ given by the number of departments at the research institute in this dataset. A likely reason for this ground-truth inflation is that certain departments, Astrophysics, Theoretical Physics and Mathematics for example, may collaborate to such an extent that their division in name may not be reflected in terms of node connection structure.

\begin{figure}[htb!]
	\centering
	\includegraphics[trim=0cm 0cm 0.0cm 0.0cm, clip, width=1.0\linewidth]{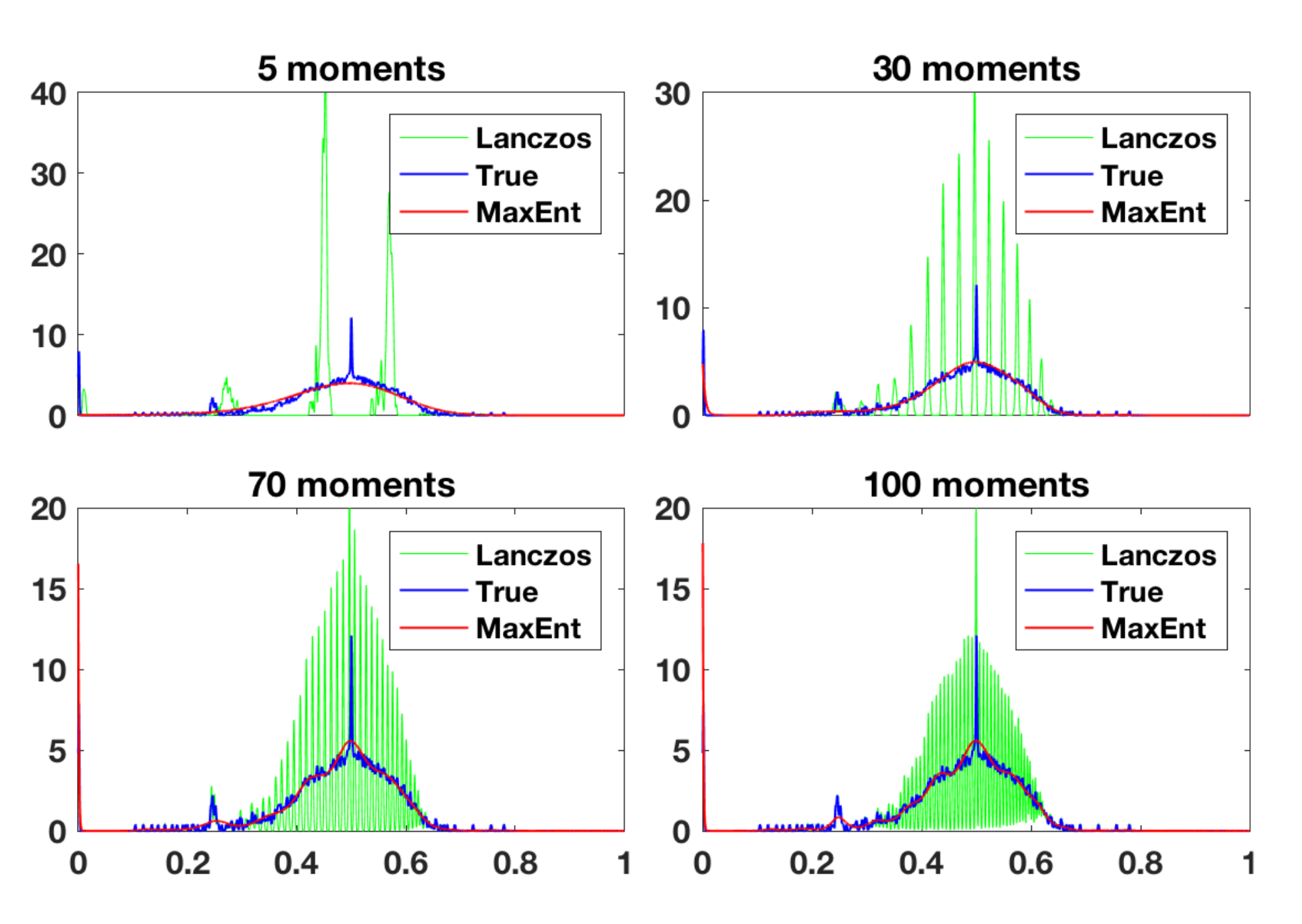}
	\caption{Spectral density for varying number of moments $m$, for both the MaxEnt and Lancsoz algorithms as well as the ground-truth.}
	\label{fig:emaildensity}	
\end{figure} 

We display the process of spectral learning for both MaxEnt and Lancsoz, by plotting the spectral density of both methods against the ground-truth in FIG \ref{fig:emaildensity}. In order to make a valid comparison, we smooth the implied density using a Gaussian kernel with $\sigma = 10^{-3}$. Whilst this number could in theory be optimised over, we considered a range of values and took the smallest for which the density was sufficiently smooth, i.e., everywhere positive on the bounded domain $[0,1]$. We note that both MaxEnt and Lancsoz approximate the ground-truth better with a greater number of moments $m$ and that Lancsoz learns the extrema of the spectrum before the bulk of the distribution while MaxEnt spectruam captures the bulk right from the start.

\begin{figure}[t]
	\centering
	\begin{subfigure}{0.5\linewidth}
		\centering
	\includegraphics[trim=0cm 0cm 0.1cm 0.0cm, clip,width=1.0\linewidth]{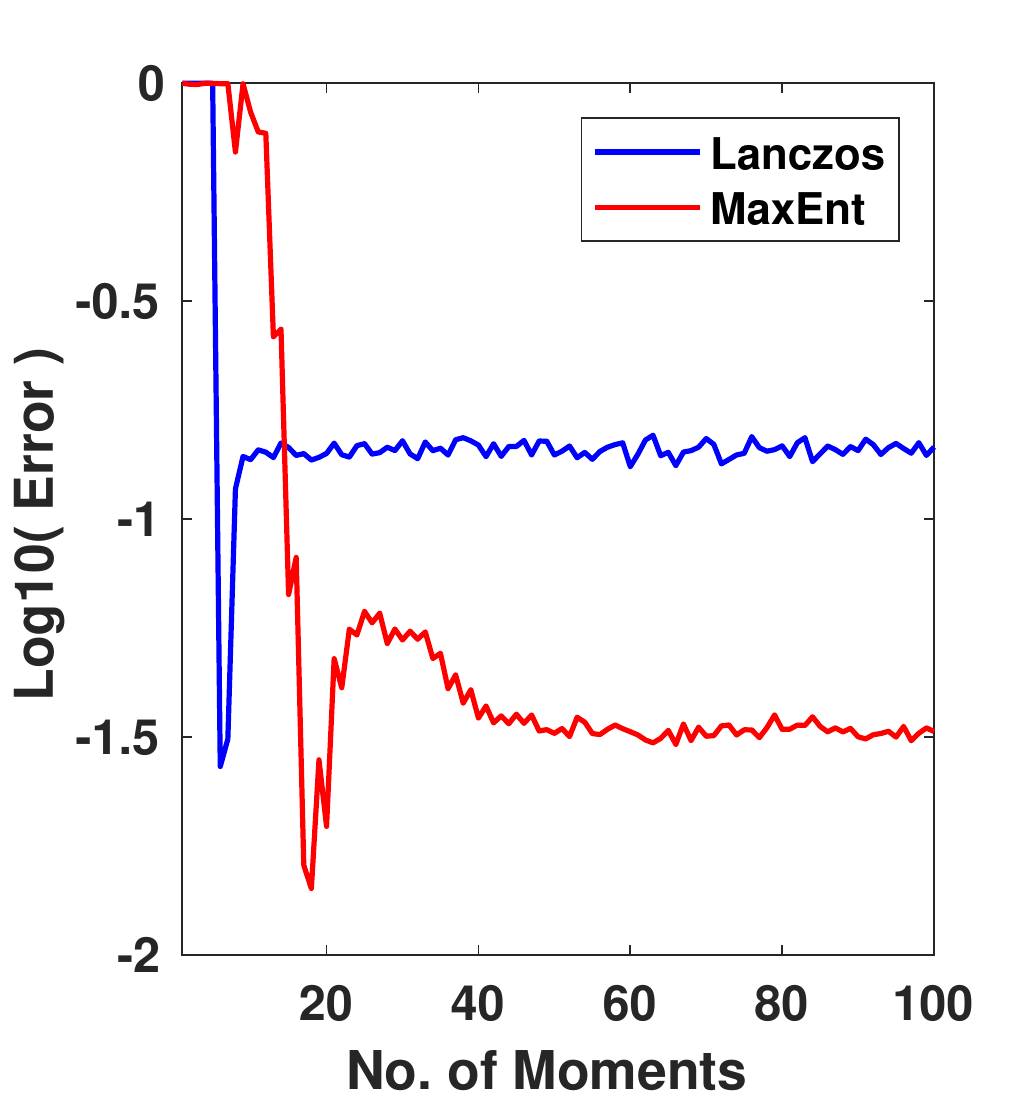}
    \caption{Email dataset}
    \label{fig:emailerror}	
	\end{subfigure}%
	\begin{subfigure}{0.5\linewidth}
		\centering
		\includegraphics[trim=0cm 0cm 0.1cm 0.0cm, clip,width=1.0\linewidth]{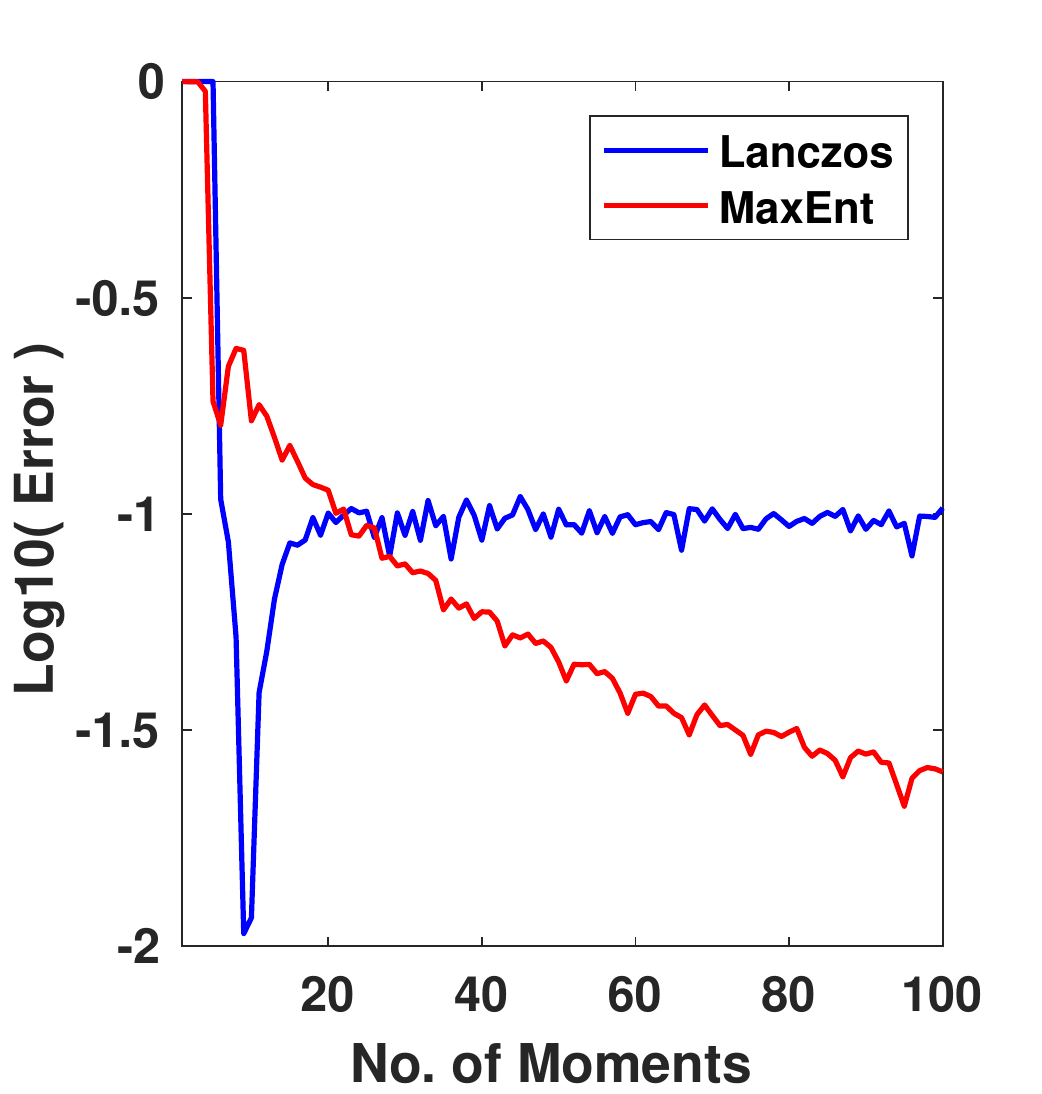}
		\caption{NetScience dataset}
		\label{fig:netscienceerror}
	\end{subfigure}
	\caption{Log error of cluster number detection using MaxEnt and Lancsoz algorithms on 2 small-scale real world networks for differing number of moments $m$.}
	\label{fig:test}
\end{figure}

We plot the log error against the number of moments for both MaxEnt and Lancsoz in FIG \ref{fig:emailerror}, with MaxEnt showing superior performance. We repeat the experiment on the Net Science collaboration network, which represents a co-authorship network of $1,589$ scientists ($n = 1,589$) working on network theory and experiment \citep{newman2006finding}. The results in FIG \ref{fig:netscienceerror} show that MaxEnt quickly outperforms the Lanczos algorithm after around $20$ moments.

\subsubsection{Large real world networks} \label{subsubsec:large_real_network}

For large datasets with $n \gg 10^{4}$, where the Cholesky decomposition becomes completely prohibitive even for powerful machines, we can no longer define a ground-truth using a complete eigen-decomposition. Alternative ``ground-truths'' supplied in \cite{mislove-2007-socialnetworks}, regarding each set of connected components with more than 3 nodes as a community, are not universally accepted. This definition, along with that of self-declared group membership \cite{yang2015defining}, often leads to contradictions with our definition of a community. A notable example is the Orkut dataset, where the number of stated communities is greater than the number of nodes \cite{snapnets}. Beyond being impossible to learn such a value from the eigenspectra, if the main reason to learn about clusters is to partition groups and to summarise networks into smaller substructures, such a definition is undesireable. 

\begin{figure}[H]
	\centering
	\begin{subfigure}{0.9\linewidth}
		\centering
	    \includegraphics[trim=0.5cm 0.5cm 0.5cm 0.8cm, clip, width=1.0\linewidth]{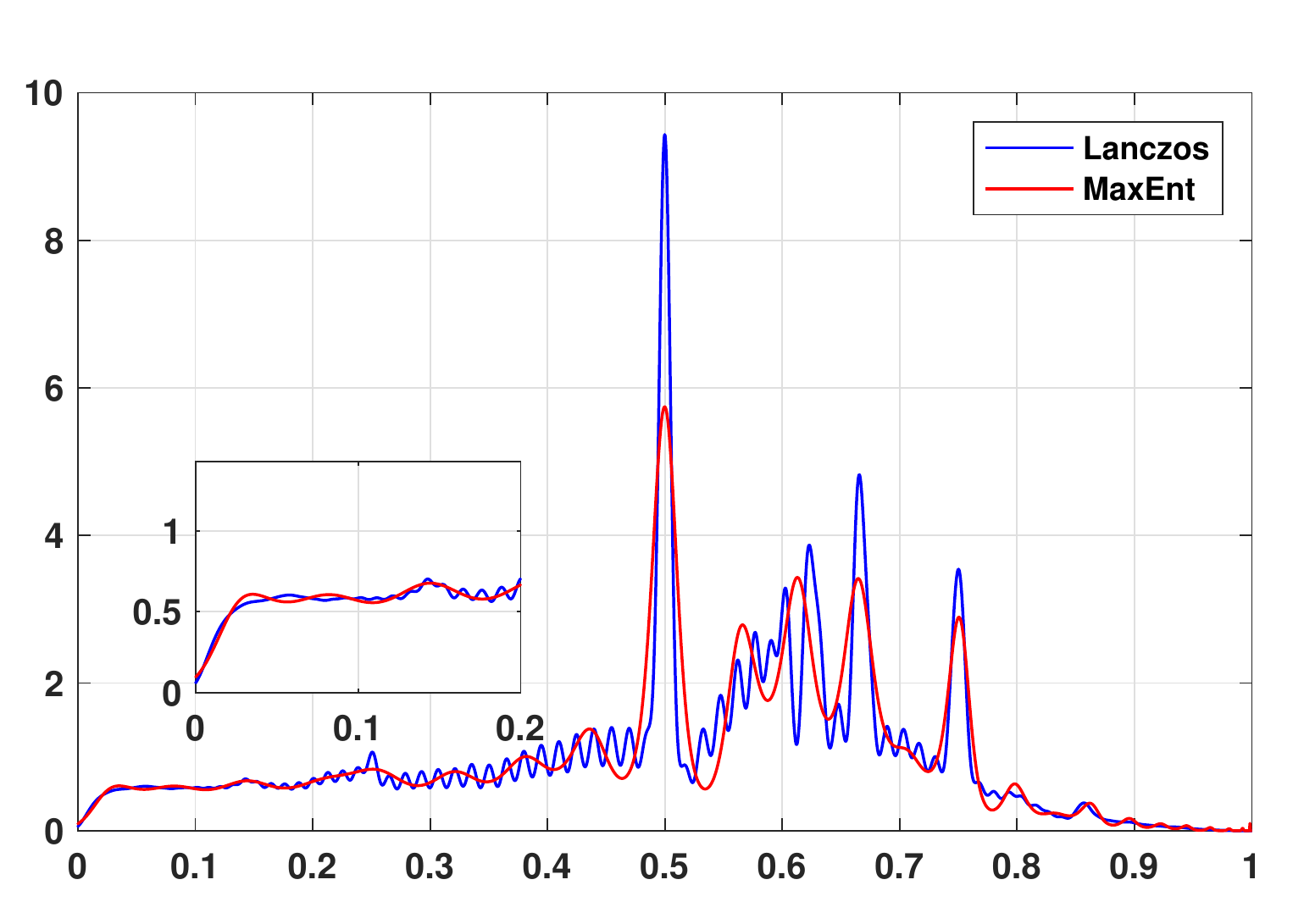}
	    \caption{DBLP}
	    \label{fig:DBLP100moments}	
	\end{subfigure}%
	
	\begin{subfigure}{0.9\linewidth}
		\centering
	    \includegraphics[trim=0.5cm 0.5cm 0.5cm 0.8cm, clip, width=1.0\linewidth]{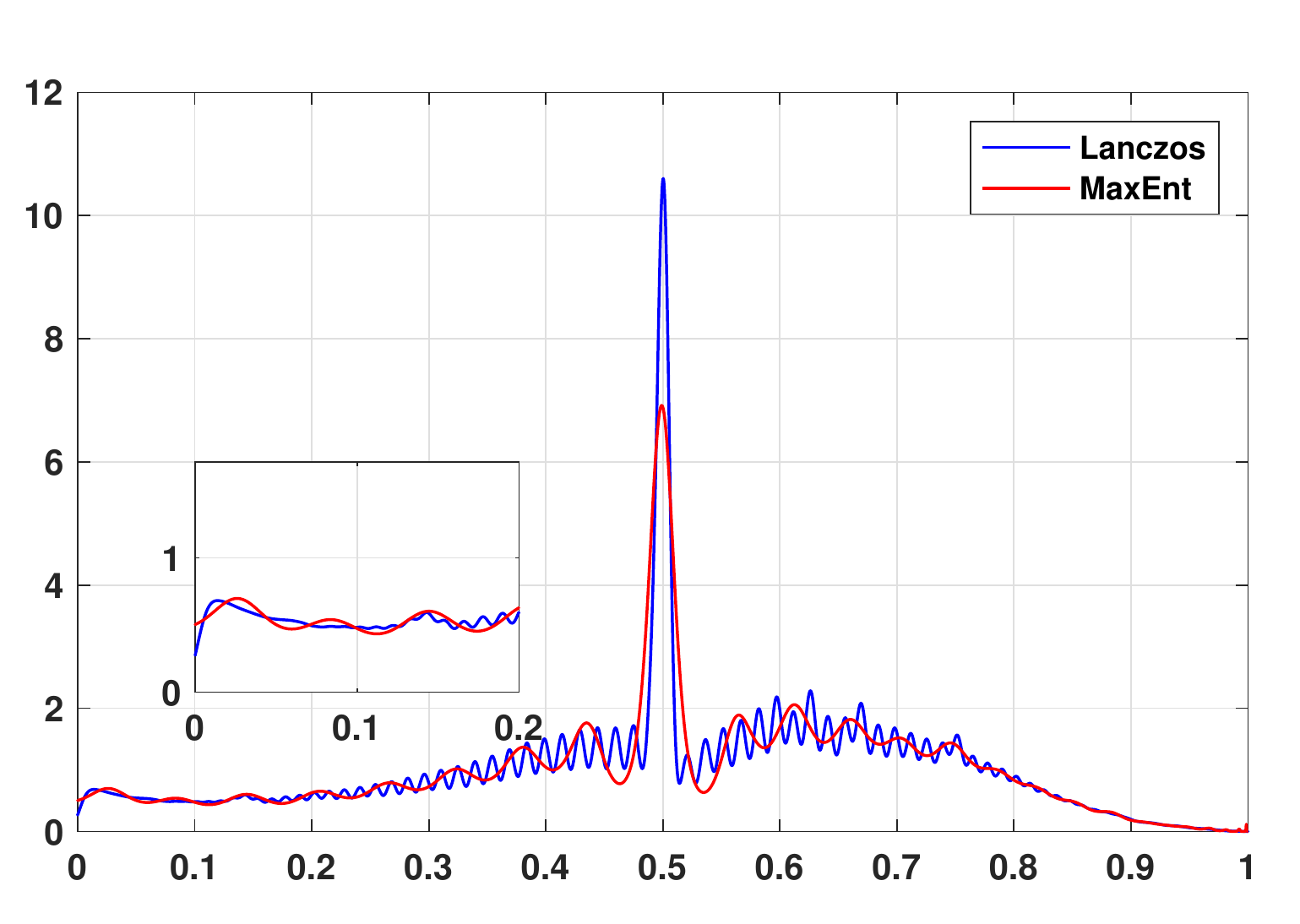}
	    \caption{Amazon}
    	\label{fig:amazon100moments}
    \end{subfigure} 

	\begin{subfigure}{0.9\linewidth}
    	\includegraphics[trim=0.5cm 0.5cm 0.5cm 0.8cm, clip, width=1.0\linewidth]{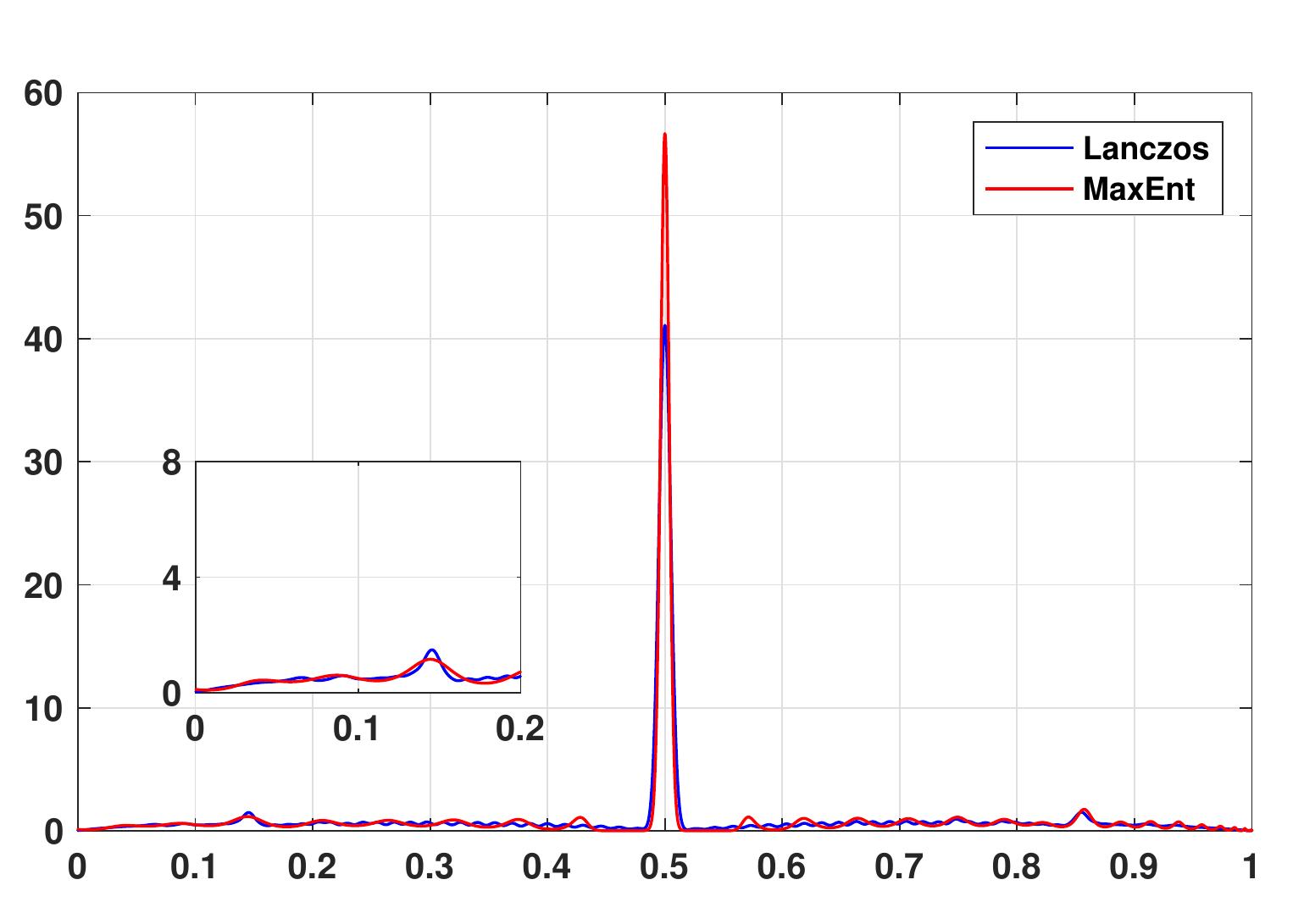}
    	\caption{YouTube}
    	\label{fig:youtube100moments}	
	\end{subfigure}%
	\caption{Spectral density for DBLP, Amazon and Youtube datasets produced by MaxEnt and Lanczos approximations (both using $m=100$).}\label{fig:bigdata}
\end{figure}

We present our findings for the number of clusters in the DBLP ($n=317,080$), Amazon ($n=334,863$) and YouTube ($n=1,134,890$) networks \cite{snapnets} in Table \ref{table:largedata}, where we use a varying number of moments. We see that for both the DBLP and Amazon networks, the number of clusters $N_{c}$ seems to converge with increasing moments number $m$, whereas for YouTube such a trend is not visible. This can be explained by looking at the approximate spectral density of the networks implied by maximum entropy in FIG \ref{fig:bigdata}. For both DBLP and Amazon (FIG \ref{fig:DBLP100moments} and \ref{fig:amazon100moments} respectively), we see that our method implies a clear spectral gap near the origin, indicating the presence of clusters. Whereas for the YouTube dataset, shown in FIG \ref{fig:youtube100moments}, no such clear spectral gap is visible and hence the number of clusters cannot be estimated accurately. 

\begin{table}[h]
	\caption{Cluster number detection by MaxEnt for DBLP ($n=317,080$), Amazon ($n=334,863$) and YouTube ($n=1,134,890$). }	\label{table:largedata}
	\begin{center}
		\begin{small}
			\begin{sc}
				\begin{tabular}{lccr}
				\hline
					\toprule
					Moments  & 40  & 70  & 100 \\
					\midrule
					DBLP    & $ 2.215\times 10^{4}$   & $8.468 \times 10^{3}$  & $8.313\times 10^{3}$   \\
					Amazon & $2.351\times 10^{4}$   & $1.146\times 10^{4}$   & $1.201\times 10^{4}$   \\
					Youtube & $4.023\times 10^{3}$   & $1.306\times 10^{4}$   & $1.900\times 10^{4}$   \\
					\bottomrule
				\end{tabular}
			\end{sc}
		\end{small}
	\end{center}
	\vskip -0.1in
\end{table}

\section{Conclusion}
In this paper, we propose a novel, efficient framework for learning a continuous approximation to the spectrum of large scale graphs, which overcomes limitations introduced from kernel smoothing. We motivate the informativeness of spectral moments using the link between random graph models and random matrix theory. We show that our algorithm is able to learn the limiting spectral densities of random graph models for which analytical solutions are known. 

We showcase the strength of this framework in two real world applications, namely, computing the similarity between different graphs and detecting the number of clusters in the graph. Interestingly, we are able to classify different real world networks with respect to their similarity to classical random graph models.

In future work, one could look at the temporal evolution of real world networks and a more complete analysis of the effect of the number of moments $m$ on the accuracy of network classification. 

\bibliographystyle{abbrvnat}
\bibliography{bibi}

\newpage

\setcounter{section}{0}
\section*{Appendix}

\section{Comment on Lanczoz algorithm}\label{subsec: lanczos}
In the state-of-the-art iterative algorithm Lanczos \cite{ubaru2017fast}, the tri-diagonal matrix  $\mvec{T}^{m\times m}$ can be derived from the moment matrix $\mvec{M}^{m\times m}$, corresponding to the discrete measure $d\alpha(\lambda)$ satisfying the moments $\mu_{i} = v^{T}X^{i}v = \int \lambda^{i}d\alpha(\lambda)$ for all $i \leq m$ \cite{golub1994matrices} and hence it can be seen as a weighted Dirac approximation to the spectral density matching the first $m$ moments.
The weight given on every Ritz eigenvalue $\lambda''_{i}$ (the eigenvalues of the matrix $\mvec{T}^{m\times m}$) is the square of the first component of the corresponding eigenvector, i.e., ${[\phi_{i}]_1}^{2}$, hence the approximated spectral density can be written as,
\begin{equation}
    \frac{1}{n}\sum_{i}^{n}\delta(\lambda-\lambda_{i}) \approx \sum_{i}^{m}w_{i}\delta(\lambda-\lambda''_{i}) =  \sum_{i}^{m}\phi_{i}[1]^{2}\delta(\lambda-\lambda''_{i}).
\end{equation}

\section{Experimental details} \label{subsec:implementation_details}
We use $d=100$ Gaussian random vectors for our stochastic trace estimation, for both MaxEnt and Lanczos \cite{ubaru2017fast}. We explain the procedure of going from adjacency matrix to Laplacian moments in Algorithm \ref{alg:preprocessing}. When comparing MaxEnt with Lanczos, we set the number of moments $m$ equal to the number of Lanczos steps, as they are both matrix vector multiplications in the Krylov subspace. We implement a quadrature MaxEnt method in Algorithm \ref{alg:maxent}. We use a grid size of $10^{-4}$ over the interval $[0,1]$ and add diagonal noise on the Hessian to improve conditioning and symmetrise it. We further use Chebyshev polynomial input instead of power moments for improved performance and conditioning. In order to normalise the moment input we use the normalised Laplacian with eigenvalues bounded by $[0,2]$ and divide by $2$. We use Python's Scipy implementation of the Newton conjugate gradient algorithm \cite{scipypackage} for the MaxEnt Lagrange multipliers. 
To make a fair comparison we take the output from Lanczos \cite{ubaru2017fast} and apply kernel smoothing \cite{lin2016approximating} before applying our cluster number estimator.

\begin{figure}[t]
    \centering
    \includegraphics[width = 1.0\linewidth]{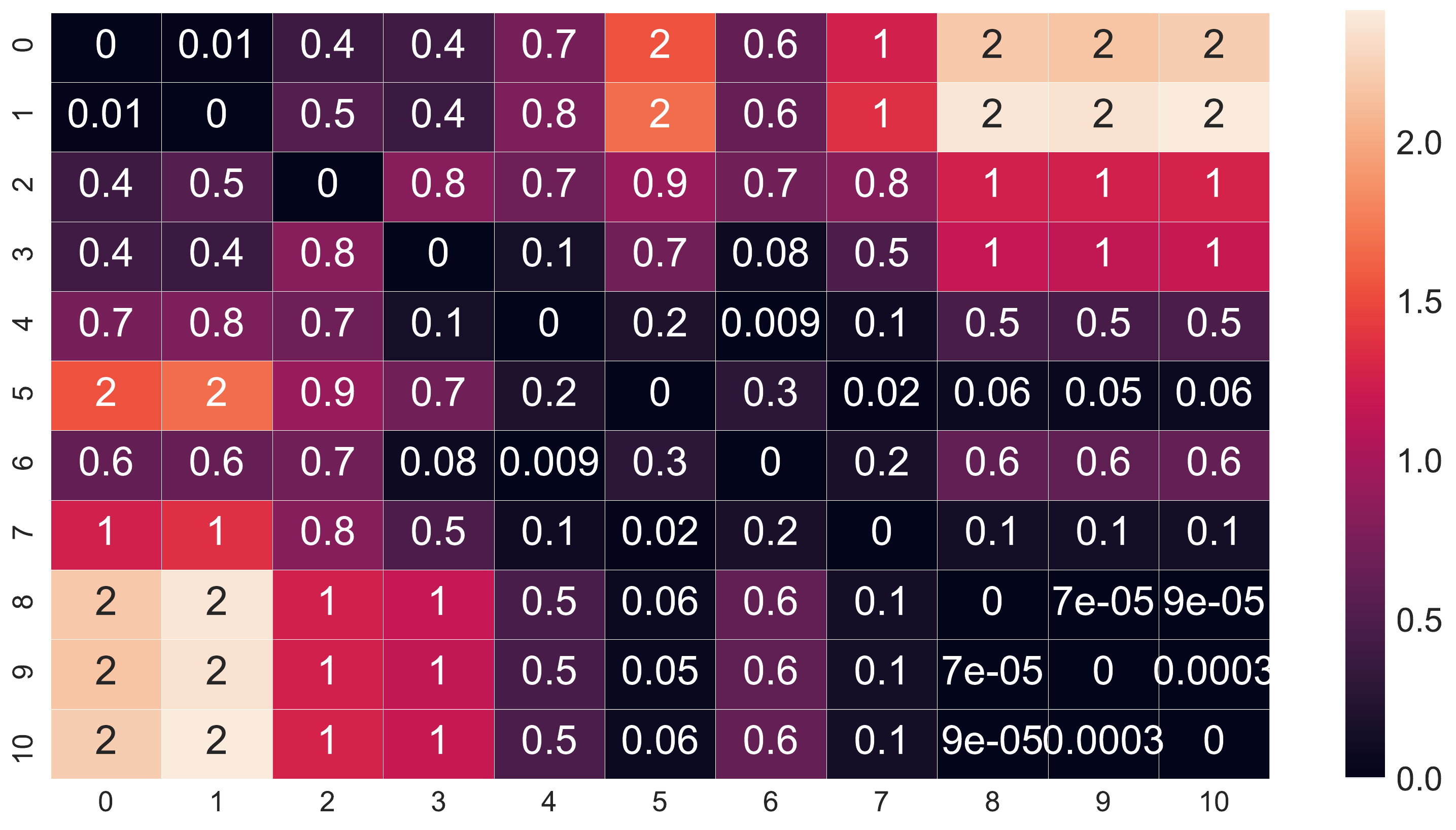}
    \caption{Symmetric KL heatmap, obtained using only $3$ moments, i.e., Gaussian approximation, between 9 graphs from the SNAP dataset, in ascending order [bio-human-gene1,
bio-human-gene2,
bio-mouse-gene,
ca-AstroPh,
ca-CondMat,
ca-GrQc,
ca-HepPh,
ca-HepTh,
roadNet-CA,
roadNet-PA,
roadNet-TX].}
    \label{fig:graphsinthewild3moments}
\end{figure}

\begin{figure}[t]
    \centering
    \includegraphics[width = 1.0\linewidth]{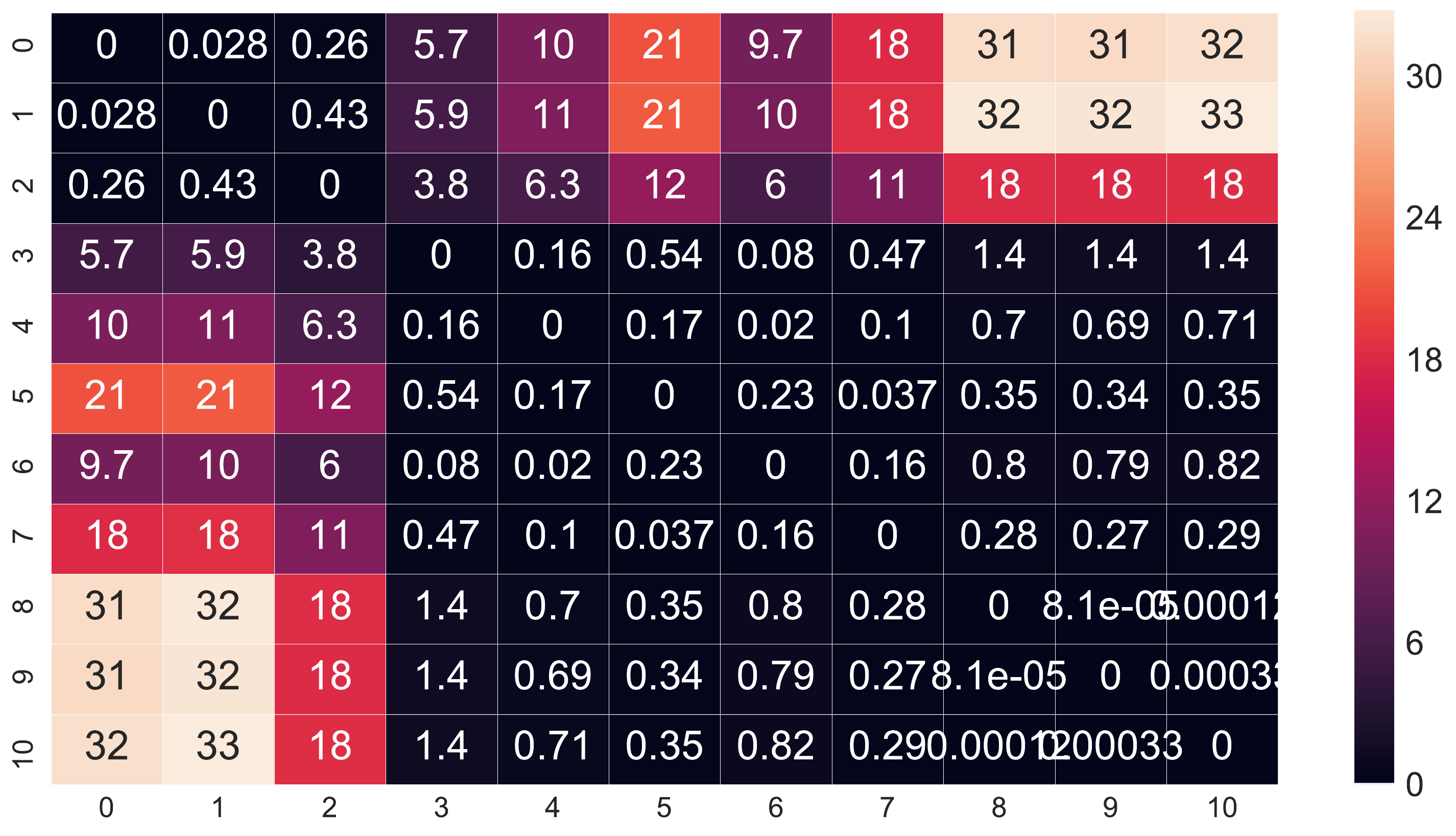}
    \caption{Symmetric KL heatmap, obtained using only $8$ moments, between 9 graphs from the SNAP dataset, in ascending order [bio-human-gene1,
bio-human-gene2,
bio-mouse-gene,
ca-AstroPh,
ca-CondMat,
ca-GrQc,
ca-HepPh,
ca-HepTh,
roadNet-CA,
roadNet-PA,
roadNet-TX].}
    \label{fig:graphsinthewild8moments}
\end{figure}

\section{MESDs of real world networks with varying number of moments}
In order to more clearly showcase the practical value of having a MESD based on a large number of moments, we show the symmetric KL divergence between real world networks using a $3$ moment Gaussian approximation. The Gaussian is fully defined by its normalization constant, mean and variance and so can be specified with $m=3$ Lagrange multipliers. The results for the same analysis as in FIG \ref{fig:graphsinthewild}, but now obtained using a $3$ moment Gaussian approximation, are shown in FIG \ref{fig:graphsinthewild3moments}. The networks are still somewhat distinguished; however, one can see for example that citation networks and road networks are less clearly distinguished to the point that inter-class distance is lessened compared to intra-class distance, which for the purpose of network classification is not a particularly helpful property. 
The problem still persists for more moments; for example, when we choose $m=8$, which is what has been reported stable for other off-the-shelf maximum entropy algorithms, 
similar results are observed in FIG \ref{fig:graphsinthewild8moments}.
In comparison, this is not the case for more moments in FIG \ref{fig:graphsinthewild} in the main text.

\end{document}